\documentclass{article}

\usepackage[table]{xcolor}

\usepackage{microtype}

\usepackage[nohyperref,accepted]{mlsys2022}

\mlsystitlerunning{Aspis: Robust Detection for Distributed Learning}

\usepackage{calc}

\usepackage{hyperref} 
\usepackage{graphics,color,wrapfig}
\usepackage{url}

\usepackage{breakurl}
\usepackage{color}
\usepackage{amsthm}
\usepackage{amsmath,amssymb}
\usepackage{amsfonts}
\usepackage{array}
\newcolumntype{P}[1]{>{\centering\arraybackslash}m{#1}}
\usepackage{multirow}
\usepackage{graphicx}
\usepackage{standalone}
\usepackage{booktabs}
\usepackage{algpseudocode} 
\usepackage{algorithm2e} 
\let\oldnl\nl
\newcommand{\nonl}{\renewcommand{\nl}{\let\nl\oldnl}}
\usepackage{makecell}
\usepackage[bottom]{footmisc}
\usepackage{enumitem}
\theoremstyle{definition}
\usepackage{tikz}

\usepackage[justification=centering]{caption} 
\usepackage{subcaption}

\usepackage{mathtools}

\usepackage{bbm}

\makeatletter
\renewcommand{\@algocf@capt@plain}{above}
\makeatother
\RestyleAlgo{ruled}
\LinesNumbered

\renewcommand{\em}{\it}

\newcommand{\calN}{\mathcal{N}}

\newcommand{\calG}{\mathcal{G}}

\newcommand{\calU}{\mathcal{U}}

\newtheorem{theorem}{Theorem}

\newtheorem{example}{Example}

\newtheorem{remark}{Remark}

{\begin{list}{}%
		{\setlength{\leftmargin}{#1}}%
		\item[]%
	}
	{\end{list}}

\newcounter{relctr} 
\everydisplay\expandafter{\the\everydisplay\setcounter{relctr}{0}} 

\AtBeginDocument{} 

\DeclareCaptionSubType * [alph]{table}
\def\*#1{\boldsymbol{\mathbf{#1}}}

\usepackage{pgf}
\usepackage{collcell}

\newcommand*{\MinNumber}{0}%
\newcommand*{\MaxNumber}{1}%

\newcommand{\ApplyGradient}[1]{%
	\pgfmathsetmacro{\PercentColor}{100.0*(#1-\MinNumber)/(\MaxNumber-\MinNumber)}%
	\edef\x{\noexpand\cellcolor{red!\PercentColor}}\x\textcolor{black}{#1}%
}
\newcolumntype{R}{>{\collectcell\ApplyGradient}{r}<{\endcollectcell}}

\begin{document}

\twocolumn[
\mlsystitle{Aspis: Robust Detection for Distributed Learning}



\mlsyssetsymbol{equal}{*}

\begin{mlsysauthorlist}
\mlsysauthor{Konstantinos Konstantinidis}{isu_ecpe}
\mlsysauthor{Aditya Ramamoorthy}{isu_ecpe}
\end{mlsysauthorlist}

\mlsysaffiliation{isu_ecpe}{Department of Electrical and Computer Engineering, Iowa State University, Ames, IA 50011 USA}

\mlsyscorrespondingauthor{Konstantinos Konstantinidis}{kostas@iastate.edu}
\mlsyscorrespondingauthor{Aditya Ramamoorthy}{adityar@iastate.edu}

\mlsyskeywords{TBD}

\vskip 0.3in

\begin{abstract}
State-of-the-art machine learning models are routinely trained on large-scale distributed clusters. Crucially, such systems can be compromised when some of the computing devices exhibit abnormal (\emph{Byzantine}) behavior and return arbitrary results to the \emph{parameter server} (PS). This behavior may be attributed to a plethora of reasons, including system failures and orchestrated attacks. Existing work suggests \emph{robust aggregation} and/or computational \emph{redundancy} to alleviate the effect of distorted gradients. However, most of these schemes are ineffective when an adversary knows the task assignment and can choose the attacked workers judiciously to induce maximal damage. Our proposed method {\it Aspis} assigns gradient computations to worker nodes using a subset-based assignment which allows for multiple consistency checks on the behavior of a worker node. Examination of the calculated gradients and post-processing (clique-finding in an appropriately constructed graph) by the central node allows for efficient detection and subsequent exclusion of adversaries from the training process. We prove the Byzantine resilience and detection guarantees of Aspis under weak and strong attacks and extensively evaluate the system on various large-scale training scenarios. The principal metric for our experiments is the test accuracy, for which we demonstrate a significant improvement of about 30\% compared to many state-of-the-art approaches on the CIFAR-10 dataset. The corresponding reduction of the fraction of corrupted gradients ranges from 16\% to 99\%.
\end{abstract}
]



\printAffiliationsAndNotice{}  

\section{Introduction}
The increased sizes of datasets and associated model complexities have established distributed training setups as the de facto method for training models at scale. A typical setup consists of one central machine (\emph{parameter server} or PS) and multiple worker machines. The PS coordinates the protocol by communicating parameters and maintaining the global copy of the model. The workers compute gradients of the loss function with respect to the optimization parameters and transmit them to the PS. The PS then updates the model. This is an iterative process and is repeated until convergence. Several frameworks including MXNet \cite{mxnet}, CNTK \cite{CNTK} and PyTorch \cite{pytorch} support this architecture.

Despite the speedup benefits of such distributed settings, they are prone to so-called \emph{Byzantine} failures, i.e., when a set of workers return malicious or erroneous computations. Faulty workers may modify their portion of the data and/or models arbitrarily. This can happen on purpose due to adversarial attacks or inadvertently due to hardware or software failures, such as bit-flipping in the memory or communication media. For example, \cite{flipping_bits_kim} showed that bit-flips in commodity DRAM can happen merely through frequent data access of the same address. Reference \cite{bit_flip_attack_rakin} exposes the vulnerability of neural networks to such failures and identifies weight parameters that could maximize accuracy degradation. As a result, the distorted gradients can derail the optimization and lead to low test accuracy. Devising training algorithms that are resilient to such failures and which can efficiently \emph{aggregate} the received gradients has inspired a series of works \cite{gupta_allerton_2019, alistarh_neurips_2018, yudong_lilisu, blanchard_krum}.


In order to alleviate the malicious effects, some existing papers use majority voting and median-based defenses \cite{aggregathor, ramchandran_saddle_point, ramchandran_optimal_rates, cong_generalized_sgd, blanchard_krum, yudong_lilisu}. Other works replicate the gradient tasks across the cluster \cite{byzshield, lagrange_cdc, detox, draco, data_encoding}. Finally, some schemes try to rank and/or detect the Byzantines \cite{regatti2020bygars, zeno, draco}.

\subsection{Contributions}

In this work, we present \emph{Aspis}, our detection-based Byzantine resilience scheme for distributed training.
Aspis uses a combination of {\it redundancy} and {\it robust aggregation}. Unlike previous methods, the redundant subset-based assignment for gradient computations is judiciously chosen such that the PS can perform global {\it consistency checks} on the behavior of the workers by examining the returned gradients. By performing clique-finding in appropriate graphs, the PS can perform \emph{detection} as a first line of defense to exclude adversaries from further being considered during the aggregation. Note that we make no assumption of privacy and our work as well as compared methods do not apply to federated learning.

Aspis has the following salient features: (i) Under weak attacks where the Byzantines act independently (without significant collusion) they will always be detected by the proposed novel clique-based algorithm. 
(ii) Furthermore, Aspis is resilient to stronger attacks where adversaries collude in optimal ways; these are much stronger than those considered in prior work. In particular, instead of simulating a random set of adversaries \cite{detox,draco}, we have crafted a non-trivial attack designed explicitly for our system such that the adversaries can evade detection and potentially corrupt more gradient values. 

For both weak and strong attacks we provide theoretical guarantees on the fraction of corrupted gradients for Aspis. Comparisons with other methods indicate reductions in the fraction of corrupted gradients ranging from 16\% to 99\%.



Finally, we present exhaustive top-1 classification accuracy results on the CIFAR-10 dataset for a variety of gradient distortion attacks coupled with choice/behavior patterns of the adversarial nodes. Our results indicate an average 30\% accuracy increase on CIFAR-10 \cite{cifar10} under the most sophisticated attacks. In summary, the performance gap between Aspis and other methods is especially stark in the strong attack scenario.


\subsection{Related Work}
Existing papers consider a wide range of assumptions regarding the maximum number of adversaries, their ability to collude, their possession of knowledge involving the data assignment and the defense protocol, and whether the adversarial machines are chosen at random or systematically. In this work, we will assume strong adversarial models as in prior work \cite{byzshield, alie, cong_generalized_sgd, ramchandran_optimal_rates, bulyan, auror}.

The first category of related prior work is called \emph{robust aggregation} 
\cite{aggregathor, ramchandran_saddle_point, ramchandran_optimal_rates, cong_generalized_sgd, blanchard_krum, yudong_lilisu}. These methods provide robustness guarantees up to a constant fraction of the nodes being adversarial. However, this fraction is usually very small and the guarantees are limited (\emph{e.g.,} only guaranteeing that the output of the aggregator has positive inner product with the true gradient \cite{bulyan, blanchard_krum}), which compromises their practicality. Also, they require significant asymptotic complexity \cite{cong_generalized_sgd} and strict assumptions such as convexity of the loss function that need to be adjusted for each individual training algorithm. Some popular robust aggregators are based on \emph{trimmed mean} \cite{cong_generalized_sgd, ramchandran_optimal_rates, bulyan} and return a subset of the values which are closest to the median element-wise. \emph{Auror} in \cite{auror} runs \emph{k-means} clustering on the gradients and outputs the mean of the largest cluster. In \emph{signSGD} \cite{SIGNSGD}, workers retain only the sign information and the PS uses majority voting aggregation. \emph{Krum} in \cite{blanchard_krum} determines a single honest worker whose gradient minimizes the distance to its $k\in\mathbb{N}$ nearest neighbors. Krum may converge to an \emph{ineffectual} model in non-convex high dimensional problems and \emph{Bulyan} \cite{bulyan} is proposed as an alternative. However, Bulyan is designed to defend only up to a small fraction of corrupted workers.

The second category is based on \emph{redundancy} and it establishes resilience by assigning each gradient task to more than one node \cite{byzshield, lagrange_cdc, detox, draco, data_encoding}. Existing techniques are sometimes combined with robust aggregation \cite{detox}. Fundamentally, these methods require higher computation load per worker but they come with stronger guarantees of correcting the erroneous gradients. It is important to note that most schemes in this category can be made to fail by a powerful \emph{omniscient} adversary which can mount judicious attacks \cite{byzshield}. \emph{DRACO} \cite{draco} builds on \cite{dimakis_cyclic_mds, tandon_gradient} and uses majority vote and Fourier decoders to filter out the adversarial effects. With $q$ Byzantines, it guarantees recovery as if the system had no adversaries, when each task is replicated $r \geq 2q+1$ times; it is not applicable if this bound is violated. Nonetheless, this requirement is very restrictive for the common assumption that up to half of the workers can be Byzantine. \emph{DETOX} in \cite{detox} extends DRACO and performs multiple stages of gradient filtering requiring smaller redundancy. However, its resilience guarantees depend heavily on a ``random choice'' of the adversaries. Subsequent work in \cite{byzshield} has crafted simple attacks to make this aggregator fail under a more careful choice of adversaries. \emph{ByzShield} in \cite{byzshield} borrows ideas from combinatorial design theory \cite{vanlint_wilson_2001} to redundantly assign the tasks and minimize the distortion fraction. Despite ByzShield's accuracy improvements, its authors have not designed an attack tailored to their method. In this work, we propose a worst-case attack that targets Aspis and prove that it is optimal.

A third category focuses on \emph{ranking} and/or \emph{detection} \cite{regatti2020bygars, draco, zeno}.
\emph{Zeno} in \cite{zeno} ranks each worker using a score that depends on the estimated loss and the magnitude of the update. Zeno requires strict assumptions on the smoothness of the loss function and the variance of the gradient estimates in order to tolerate an adversarial majority in the cluster. Similarly, \emph{ByGARS} \cite{regatti2020bygars} computes reputation scores for the nodes based on an auxiliary dataset; these scores are used to weigh the contribution of each gradient to the model update. Its convergence result depends on the assumption of a strongly convex loss function. In contrast, our method does not rely on such restrictive assumptions. Also, these works have not used or constructed worst-case attacks that would evaluate the methods in adversarial settings.

\begin{figure}
	\begin{center}
		\includegraphics[width=0.48\textwidth]{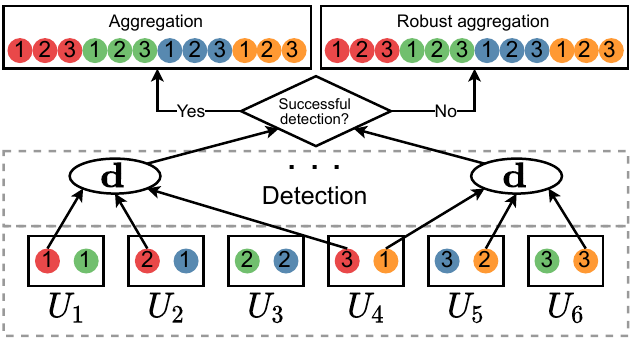}
	\end{center}
	\caption{Aggregation of gradients on a cluster.}
	\label{fig:aggregation_general}
\end{figure}

\section{Distributed Training Formulation}
\label{sec:formulation}
The formulation we discuss is standard in distributed deep learning. Assume a loss function $l_i(\mathbf{w})$ for the $i^{\mathrm{th}}$ sample where $\mathbf{w}\in\mathbb{R}^d$ is the parameter set of the model.\footnote{The paper's heavily-used notation is summarized in Appendix Table \ref{table:notation}.} We use \emph{mini-batch Stochastic Gradient Descent} (SGD) to minimize the loss over the entire dataset, i.e.,
\begin{equation*}
	\min_{\mathbf{w}}L(\mathbf{w})=\min_{\mathbf{w}}\frac{1}{n}\sum\limits_{i=1}^nl_i(\mathbf{w})
\end{equation*}
where $n$ is the dataset size. Initially, $\mathbf{w}$ is randomly set to $\mathbf{w}_0$ ($\mathbf{w}_t$ is the model state at the end of iteration $t$). A random \emph{batch} $B_t$ of $b$ samples is chosen to perform the update in the $t^{\mathrm{th}}$ iteration. Thus,
\begin{equation}
	\label{eq:vanilla_sgd_update}
	\mathbf{w}_{t+1}=\mathbf{w}_{t}-\eta_t\frac{1}{|B_t|}\sum\limits_{i\in B_t}\nabla l_i(\mathbf{w}_t)
\end{equation}
where $\eta_t$ is the learning rate of the $t^{\mathrm{th}}$ iteration. The workers, denoted $U_1, U_2, \dots, U_{K}$, compute gradients on subsets of the batch. The training is \emph{synchronous}, i.e., the PS waits for all workers to return before performing an update. It stores the dataset and the model and coordinates the protocol.

\textbf{Task assignment}: Each batch $B_t$ is split into $f$ disjoint files $\{B_{t,i}\}_{i=0}^{f-1}$. These files are then assigned to the workers according to our placement policy. Computational redundancy is introduced by assigning a given file to $r>1$ workers; we will occasionally use the term \emph{group} (of the assigned workers) to refer to a file. It follows that each worker is responsible for $l = fr/K$ files ($l$ is the \emph{computation load}). We let $\calN(U_j)$ to be the set of files assigned to worker $U_j$ and $\calN(B_{t,i})$ to be the group of workers that are assigned file $B_{t,i}$. Our placement scheme is such that $\calN(B_{t,i})$ uniquely identifies the file $B_{t,i}$; thus, we will sometimes refer to the file $B_{t,i}$ by its worker assignment, $\calN(B_{t,i})$. The actual placement algorithm will be presented in Section \ref{sec:task_assignment}.

\textbf{Adversary model}: We assume that at most $q$ workers can be adversarial, i.e., they can return arbitrary values to the PS. The workers know the data assignment of all nodes, the parameters $\mathbf{w}_t$ and the defense at {\it every} iteration (\emph{omniscient} attack); they can also collude. The adversarial machines may change at every single iteration. We will suppose that $q < K/2$. We emphasize that our attack setting is more powerful than random failures considered in related redundancy-based work \cite{detox, draco}.
For each assigned file $B_{t,i}$ a worker $U_j$ will return the value $\hat{\mathbf{g}}_{t,i}^{(j)}$ to the PS. Then,
\begin{equation}
	\hat{\mathbf{g}}_{t,i}^{(j)} = \left\{
	\begin{array}{ll}
		\mathbf{g}_{t,i} & \text{ if } U_j \text{ is honest},\\
		\mathbf{*} & \text{otherwise}, \\
	\end{array}
	\right.
\end{equation}
where $\mathbf{g}_{t,i}$ is the sum of the loss gradients on all samples in file $B_{t,i}$, i.e.,
\begin{equation*}
	\mathbf{g}_{t,i} = \sum\limits_{j\in B_{t,i}}\nabla l_j(\mathbf{w}_t)
\end{equation*}
and $\mathbf{*}$ is any arbitrary vector in $\mathbb{R}^d$. Within this setup, we examine multiple adversarial scenarios.

\textbf{Training}: We will refer to Figure \ref{fig:aggregation_general} for this exposition. There are $K=6$ machines and $f=4$ distinct files (represented by colored circles) replicated $r=3$ times.\footnote{Some arrows and ellipses have been omitted from Figure \ref{fig:aggregation_general}; however, all files will be going through detection.} Each of the workers is assigned to $l=2$ files and computes the sum of gradients (or an distorted value) on each of them. The ``\textbf{d}'' ellipses refer to detection operations the PS performs immediately after receiving all the gradients.

The algorithm starts with the assignment of files to workers. Subsequently, each worker $U_i$ will compute all $l$ file gradients that involve its assigned files $\calN(U_i)$ and return them to the PS. In every iteration, the PS will initially run our detection algorithm in an effort to identify the $q$ adversaries and will act differently depending on the detection outcome.
\begin{itemize}[wide, labelwidth=!, labelindent=0pt]
	\item \emph{Case 1: Successful detection}. The PS will ignore all detected faulty machines and keep only the gradients from the remaining workers. Assume that $h$ workers $U_{i_1}, U_{i_2}, \dots, U_{i_h}$ have been identified as honest. For each of the $f$ files if there is at least one honest worker that processed it, the PS will pick one of the ``honest'' gradient values. The chosen gradients are then averaged for the update (\emph{cf.} Eq. \eqref{eq:vanilla_sgd_update}). For instance, in Figure \ref{fig:aggregation_general}, assume that $U_1$, $U_2$ and $U_4$ have been identified as faulty. During aggregation, the PS will ignore the red file as all 3 copies have been compromised. For the orange file, it will pick either the gradient computed by $U_5$ or $U_6$ as both of them are honest.
	
	\item \emph{Case 2: Unsuccessful detection}. During aggregation, the PS will perform a majority vote across the computations of each file. Recall that each file has been processed by $r$ workers. For each such file $B_{t,i}$, the PS decides a majority value $\mathbf{m}_i$
	
	\begin{equation}
		\label{eq:basic_majority}
		\mathbf{m}_i := \mathrm{majority}\left\{\hat{\mathbf{g}}_{t,i}^{(j)}: U_j \in \calN(B_{t,i})\right\}.
	\end{equation}
	
	
	Assume that $r$ is odd and let $r'=\frac{r+1}{2}$. Under the rule in Eq. \eqref{eq:basic_majority}, the gradient on a file is distorted only if at least $r'$ of the computations are performed by Byzantines. Following the majority vote, we will further filter the gradients using coordinate-wise median and will be referring to the combination of these two steps as \emph{robust aggregation}; a similar setup was considered in \cite{byzshield, detox}. For example, in Figure \ref{fig:aggregation_general}, all returned values for the red file will be evaluated by a majority vote function on the PS which decides a single output value; a similar voting is done for the other 3 files. After the voting process, Aspis applies coordinate-wise median on the ``winning'' gradients $\mathbf{m}_i$, $i=0,1,\dots,f-1$.
\end{itemize}

\begin{algorithm}[!t]
	\KwIn{
		Dataset of $n$ samples, batchsize $b$, computation load $l$, redundancy $r$, \newline number of files $f$, maximum iterations $T$, file assignments $\{\calN(U_i)\}_{i=1}^{K}$.
	}
	{
		\abovedisplayskip=0pt
		\belowdisplayskip=0pt
		The PS randomly initializes model's parameters to $\mathbf{w}_0$.\\
		\For{$t = 1$ to $T$}{
			PS chooses a random batch $B_t\subseteq\{1,2,\dots,n\}$ of $b$ samples, partitions it into $f$ files $\{B_{t,i}\}_{i=0}^{f-1}$ and assigns them to workers according to $\{\calN(U_i)\}_{i=1}^{K}$. It then transmits $\mathbf{w}_t$ to all workers.\\
			\For{each worker $U_j$}{
				\eIf{$U_j$ is honest}{
					\For{each file $i \in \calN(U_j)$}{
						$U_j$ computes the sum of gradients $$\hat{\mathbf{g}}_{t,i}^{(j)}=\sum\limits_{k\in B_{t,i}}\nabla l_k(\mathbf{w}_t).$$
					}
				}{
					$U_j$ constructs $l$ adversarial vectors $$\hat{\mathbf{g}}_{t,i_1}^{(j)},\hat{\mathbf{g}}_{t,i_2}^{(j)},\dots,\hat{\mathbf{g}}_{t,i_l}^{(j)}.$$
				}
				$U_j$ returns $\hat{\mathbf{g}}_{t,i_1}^{(j)},\hat{\mathbf{g}}_{t,i_2}^{(j)},\dots,\hat{\mathbf{g}}_{t,i_l}^{(j)}$ to the PS.
			}
			PS runs a detection algorithm to identify the adversaries.
			
			\eIf{detection is successful}{
				Let $H$ be the detected honest workers. Initialize a non-corrupted gradient set as $\calG = \emptyset$.\\
				\For{each file in $\{B_{t,i}\}_{i=0}^{f-1}$}{
					PS chooses the gradient of a worker in $\calN(B_{t,i}) \cap H$ (if non-empty) and adds it to $\calG$.
				}
				\begin{equation*}
					\mathbf{w}_{t+1}=\mathbf{w}_{t}-\eta_t\frac{1}{|\calG|}\sum\limits_{\mathbf{g}\in \calG}\mathbf{g}.
				\end{equation*}
			}{
				\For{each file in $\{B_{t,i}\}_{i=0}^{f-1}$}{
					PS determines the $r$ workers in $\calN(B_{t,i})$ which have processed $B_{t,i}$ and computes $$\mathbf{m}_i = \mathrm{majority}\left\{\hat{\mathbf{g}}_{t,i}^{(j)}: j \in \calN(B_{t,i})\}\right\}.$$
				}
				PS updates the model via
				\begin{equation*}
					\mathbf{w}_{t+1} = \mathbf{w}_{t} - \eta_t \times \mathrm{median}\{\mathbf{m}_i: i = 0,1,\dots,f-1\}.
				\end{equation*}
			}
		}
	}
	\caption{Proposed Aspis aggregation algorithm to alleviate Byzantine effects.}
	\label{alg:main_algorithm}
\end{algorithm}

The details of this procedure are described in Algorithm \ref{alg:main_algorithm}. In both cases, we ensure that there are no floating point precision issues in our implementation, i.e., all honest workers assigned to $B_{t,i}$ will return the exact same gradient. In general, even if such issues arise they can easily be handled. In Case 1, if the honest gradients for a file were different the PS could average them.  In Case 2, we could cluster the gradients and compute the average of the largest cluster.

{\bf Metrics}: We are interested in two main metrics, the fraction of distorted files and the top-1 test accuracy of the final trained model. We evaluate these metrics for the various competing methods.

\begin{algorithm}[!t]
	\KwIn{Batch size $b$, computation load $l$, redundancy $r$ and worker set $\calU$, $|\calU| = K$.}
	{
		\abovedisplayskip=0pt
		\belowdisplayskip=0pt
		PS partitions batch $B_t$ into $f = \binom{K}{r}$ disjoint files of $b/f$ samples each $$B_t=\left\{B_{t,i}: i=0,1,\dots,f-1\right\}.$$\\
		PS constructs all subsets $S_0,S_1,\dots,S_{\binom{K}{r}-1}$ of $\calU = \{U_1,U_2,\dots,U_{K}\}$ such that $\forall i, |S_i| = r$.\\
		\For{$i = 0$ to $f-1$}{
			PS identifies all workers in group $S_i = \{U_{j_1}, U_{j_2}, \dots, U_{j_r}\}$ and assigns the file indexed with $i$ in $B_t$ to all of them. Formally, $\calN(U_j) = \calN(U_j) \cup \{B_{t,i}\}$ for $j\in\{j_1,j_2,\dots,j_r\}$.
		}
	}
	\caption{Aspis subset-based file assignment.}
	\label{alg:subset_placement}
\end{algorithm}

\section{Task Assignment}
\label{sec:task_assignment}
In this section, we propose our technique which determines the allocation of gradient tasks to workers in Aspis. Let $\calU$ be the set of workers. Our scheme has $|\calU| \leq f$ (i.e., fewer workers than files).

To allocate the batch of an iteration, $B_t$, to the $K$ workers, first, we will partition $B_t$ into $f=\binom{K}{r}$ disjoint files $B_{t,0},B_{t,1},\dots,B_{t,f-1}$; recall that $r$ is the redundancy. Following this, we associate each file with exactly one of the subsets $S_0,S_1,\dots,S_{\binom{K}{r}-1}$ of $\{U_1,U_2,\dots,U_{K}\}$ each of cardinality $r$, essentially using a bijection. Each file contains $b/f$ samples. The details are specified in Algorithm \ref{alg:subset_placement}. 
The following example showcases the proposed protocol of Algorithm \ref{alg:subset_placement}.


\begin{example}
	\label{ex:placement_K7_r3}
	Consider $K=7$ workers $U_1,U_2\dots,U_{7}$ and $r=3$. Based on our protocol, the $f=\binom{7}{3} =35$ files of each batch $B_t$ are associated one-to-one with 3-subsets of $\calU$, e.g., the subset $S_0 = \{U_1,U_2,U_3\}$ corresponds to file $B_{t,0}$ and will be processed by $U_1$, $U_2$ and $U_3$.
\end{example}

\begin{remark}
	Our task assignment ensures that every pair of workers processes $\binom{K-2}{r-2}$ files. Moreover, the number of adversaries is $q < K/2$. Thus, upon receiving the gradients from the workers, the PS can examine them for consistency and flag certain nodes as adversarial if their computed gradients differ from $q+1$ or more of the other nodes. We use this intuition to detect and mitigate the adversarial effects and to compute the fraction of corrupted files.
\end{remark}

\section{Adversarial Detection}
\label{sec:detection}

The PS will run our detection method in every iteration as our model assumes that the adversaries can be different across different steps. The following analysis applies to any iteration; thus, the iteration index $t$ will be omitted from most of the notation used in this section. Let the current set of adversaries be $A \subset \{U_1,U_2,\dots,U_K\}$ with $|A|=q$; also, let $H$ be the honest worker set. The set $A$ is unknown but our goal is to provide an estimate $\hat{A}$ of it. Ideally, the two sets should be identical. In general, depending on the adversarial behavior, we will be able provide a set $\hat{A}$ such that $\hat{A} \subseteq A$. For each file, there is a group of $r$ workers which have processed it and there are ${r\choose 2}$ pairs of workers in each group. Each such pair may or may not agree on the gradient value for the file. For an iteration, let us encode the agreement of workers $U_{j_1}, U_{j_2}$ on a common file $i$ of them as
\begin{equation}
	\alpha_i^{(j_1,j_2)} := \left\{
	\begin{array}{ll}
		1 & \text{if } \hat{\mathbf{g}}_{i}^{(j_1)} = \hat{\mathbf{g}}_{i}^{(j_2)},\\
		0 & \text{otherwise},
	\end{array}
	\right.
\end{equation}
where the iteration subscript $t$ is skipped for the computed gradients.
Then, across all files, let us denote the total number of agreements between a pair of workers $U_{j_1}, U_{j_2}$ by
\begin{equation}
	\alpha^{(j_1,j_2)} := \sum_{i\in \mathcal{N}(U_{j_1}) \cap \mathcal{N}(U_{j_2})}\alpha_i^{(j_1,j_2)}.
\end{equation}
Since the placement is known, the PS can always perform the above computation. Next, we form an undirected graph $\mathbf{G}$ whose vertices correspond to all workers $\{U_1, U_2, \dots, U_{K}\}$. An edge $(U_{j_1}, U_{j_2})$ exists in $\mathbf{G}$ only if the computed gradients of $U_{j_1}$ and $U_{j_2}$ match in all their  ${{K-2}\choose {r-2}}$ common groups.

A \emph{clique} in an undirected graph is defined as a subset of vertices in which there is an edge between any pair of them. A \emph{maximal clique} is one that cannot be enlarged by adding additional vertices to it. A \emph{maximum clique} is a clique such that there is no clique with more vertices in the given graph. We note that the set of honest workers $H$ will pair-wise agree everywhere. In particular, this implies that the subset $H$ forms a clique (of size $K-q$) within $\mathbf{G}$. The clique containing the honest workers may not be maximal. However, it will have a size at least $K-q$. Let the maximum clique on $\mathbf{G}$ be $M_{\mathbf{G}}$. Any worker $U_j$ with $\deg(U_j) < K-q-1$ will not belong to a maximum clique and can straight away be eliminated as a ``detected'' adversary.

\begin{algorithm}[!t]
	\KwIn{Computed gradients $\hat{\mathbf{g}}_{t,i}^{(j)}$, $i=0,1,\dots,f-1$, $j=1,2,\dots,K$, redundancy $r$ and empty graph $\mathbf{G}$ with worker vertices $\calU$.}
	{
		\abovedisplayskip=0pt
		\belowdisplayskip=0pt
		\For{each pair $(U_{j_1}, U_{j_2}), j_1\neq j_2$ of workers}{
			PS computes the number of agreements $\alpha^{(j_1,j_2)}$ of the pair $U_{j_1}, U_{j_2}$ on the gradient value.
			
			\If{$\alpha^{(j_1,j_2)} = {{K-2}\choose {r-2}}$}{
				Connect vertex $U_{j_1}$ to vertex $U_{j_2}$ in $\mathbf{G}$.
			}
		}
		
		PS enumerates all $k$ maximum cliques $M_{\mathbf{G}}^{(1)}, M_{\mathbf{G}}^{(2)}, \dots, M_{\mathbf{G}}^{(k)}$ in $\mathbf{G}$.
		
		\eIf{there is a unique maximum clique $M_{\mathbf{G}}$ ($k=1$)}{
			PS determines the honest workers $H = M_{\mathbf{G}}$ and the adversarial machines $\hat{A} = \mathcal{U} - M_{\mathbf{G}}$.
		}{
			PS declares unsuccessful detection.
		}
	}
	\caption{Proposed Aspis graph-based detection.}
	\label{alg:detection}
\end{algorithm}

The essential idea of our detection is to run a clique-finding algorithm on $\mathbf{G}$ (summarized in Algorithm \ref{alg:detection}). If we find a unique maximum clique, we declare it to be the set of honest workers; the gradients from the detected adversaries are ignored. On the other hand, if there is more than one maximum clique, we resort to the robust aggregation technique discussed in Section \ref{sec:formulation}. Let us denote the number of distorted tasks upon Aspis detection and aggregation by $c^{(q)}$ and its maximum value (under the worst-case attack) by $c_{\mathrm{max}}^{(q)}$. The \emph{distortion fraction} is $\epsilon := c^{(q)}/f$. Clique-finding is well-known to be an NP-complete problem \cite{karp1972}. Nevertheless, there are fast practical algorithms that have excellent performance on graphs even up to hundreds of nodes \cite{cazals_clique, tomita_clique}. Specifically, the authors of \cite{tomita_clique} have shown that their proposed algorithm which enumerates all maximal cliques has similar complexity as other methods \cite{robson_1986, tarjan_1977} which are used to find a single maximum clique. We utilize this algorithm which is proven to be optimal. Our extensive experimental evidence suggests that clique-finding is not a computation bottleneck for the size and structure of the graphs that Aspis uses. We have experimented with clique-finding on a graph of $K=100$ workers and $r=5$ for different values of $q$; in all cases, enumerating all maximal cliques took no more than 15 milliseconds. These experiments and asymptotic complexity of entire protocol are addressed in Appendix Section \ref{appendix:asymptotic}. 

\begin{figure}[t]
	\centering
	\begin{subfigure}[b]{0.43\textwidth}
		\centering
		\includegraphics[scale=0.3]{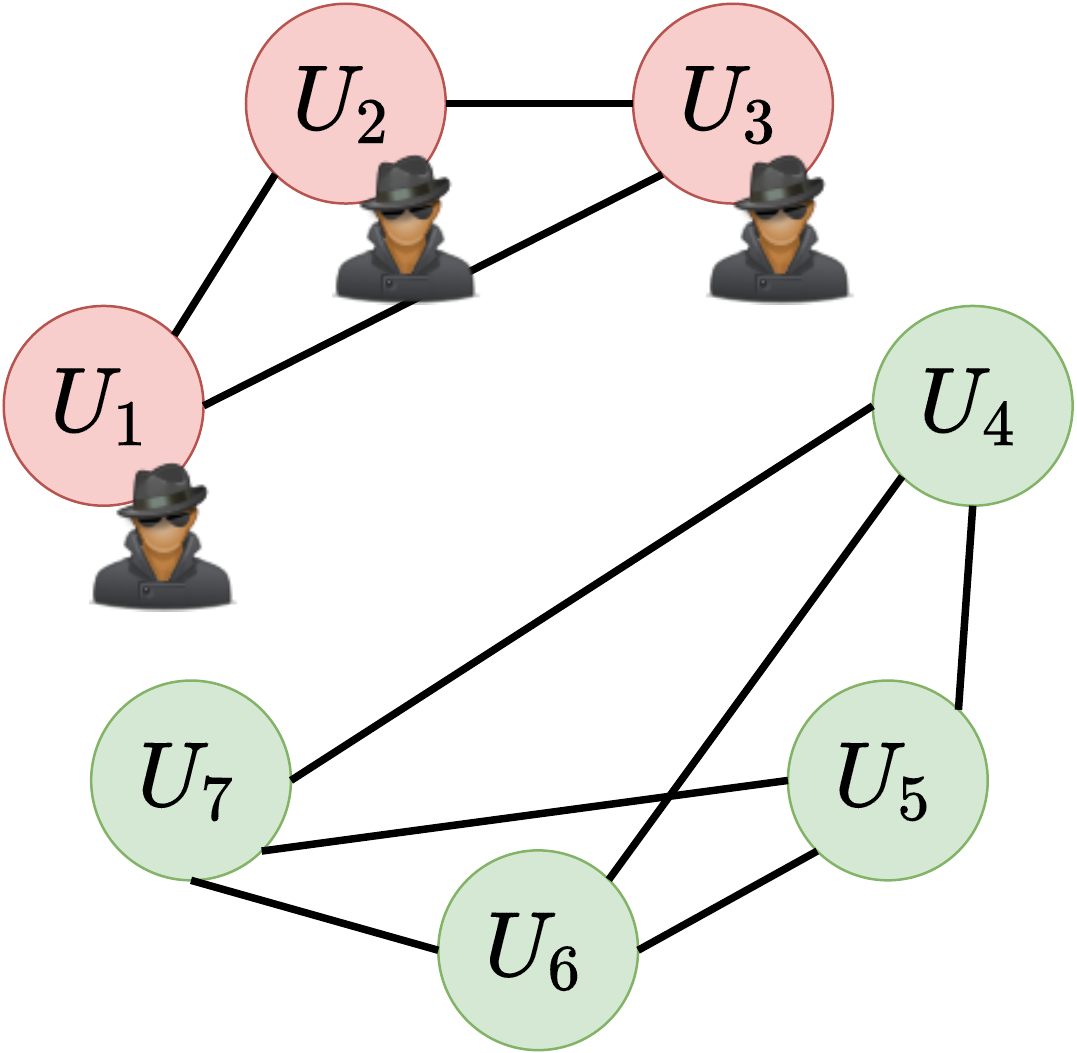}
		\caption{Unique max-clique, detection succeeds.}
		\label{fig:subset_assignment_K7_r3_graph_success}
	\end{subfigure}
	\hspace{0.01\textwidth}
	\begin{subfigure}[b]{0.43\textwidth}
		\centering
		\includegraphics[scale=0.3]{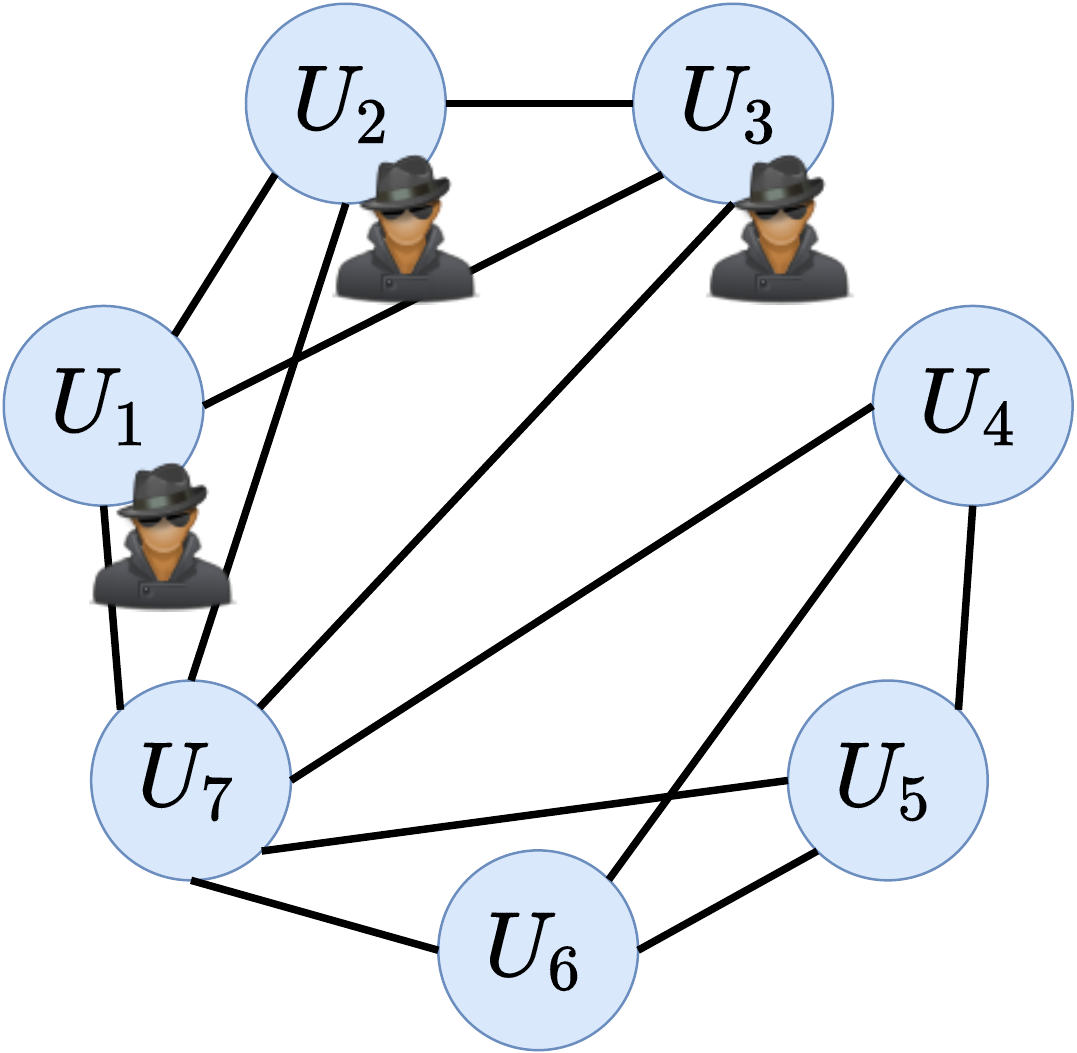}
		\caption{Two max-cliques, detection fails.}
		\label{fig:subset_assignment_K7_r3_graph_failure}
	\end{subfigure}
	\caption{Detection graph $\mathbf{G}$ for $K=7$ workers among which $U_1$, $U_2$ and $U_3$ are the adversaries.}
\end{figure}

\subsection{Weak Adversarial Strategy}
\label{sec:weak_subset_attack}
We first consider a class of weak attacks where the Byzantine nodes attempt to distort the gradient on any file that they participate in. For instance, a node may try to return arbitrary gradients on all its files. A more sophisticated strategy (but still weak) for the node would be to match the result with other adversarial nodes that process the same file. In both cases, it is clear that the Byzantine node will disagree with at least $K-q$ honest nodes and thus the degree of the node in $\mathbf{G}$ will be at most $q-1 < K-q-1$ and it will not be part of the maximum clique. Thus, each of the adversaries will be detected and their returned gradients will not be considered further. The algorithm declares the (unique) maximum clique as honest and proceeds to aggregation (\emph{cf.} Section \ref{sec:formulation}). The only files that can be distorted in this case are those that consist exclusively of adversarial nodes.

Figure \ref{fig:subset_assignment_K7_r3_graph_success} (corresponding to Example \ref{ex:placement_K7_r3}) shows an example where in a cluster of size $K=7$, the $q=3$ adversaries are $A = \{U_1, U_2, U_3\}$ and the remaining workers are honest with $H = \{U_4, U_5, U_6, U_7\}$. In this case, the unique maximum clique is $M_{\mathbf{G}} = H$ and detection is successful. Under this attack, the number of distorted tasks are those whose all copies have been compromised, i.e., $c^{(q)} = \binom{q}{r}$.

\begin{remark}
	We emphasize that even though we call this attack ``weak'' this is the attack model considered in several prior works \cite{detox, draco}. To our best knowledge, most of them have not considered the adversarial problem from the lens of detection.
\end{remark}


\begin{figure}[t]
	\centering
	\begin{subfigure}[b]{0.43\textwidth}
		\centering
		\includegraphics[scale=0.4]{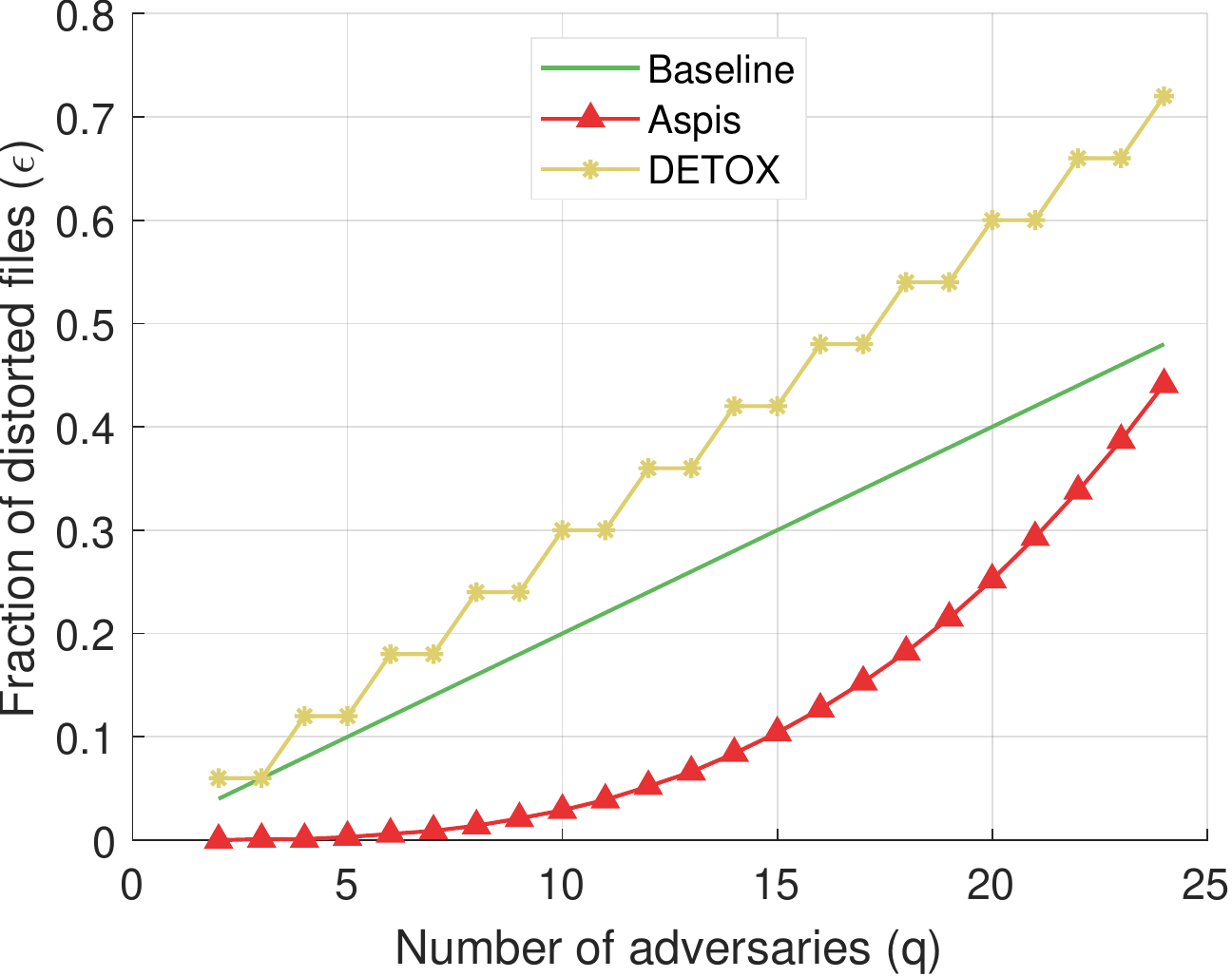}
		\caption{Optimal attacks.}
		\label{fig:epsilon_simulations_optimal_K50}
	\end{subfigure}
	\hspace{0.01\textwidth}
	\begin{subfigure}[b]{0.43\textwidth}
		\centering
		\includegraphics[scale=0.4]{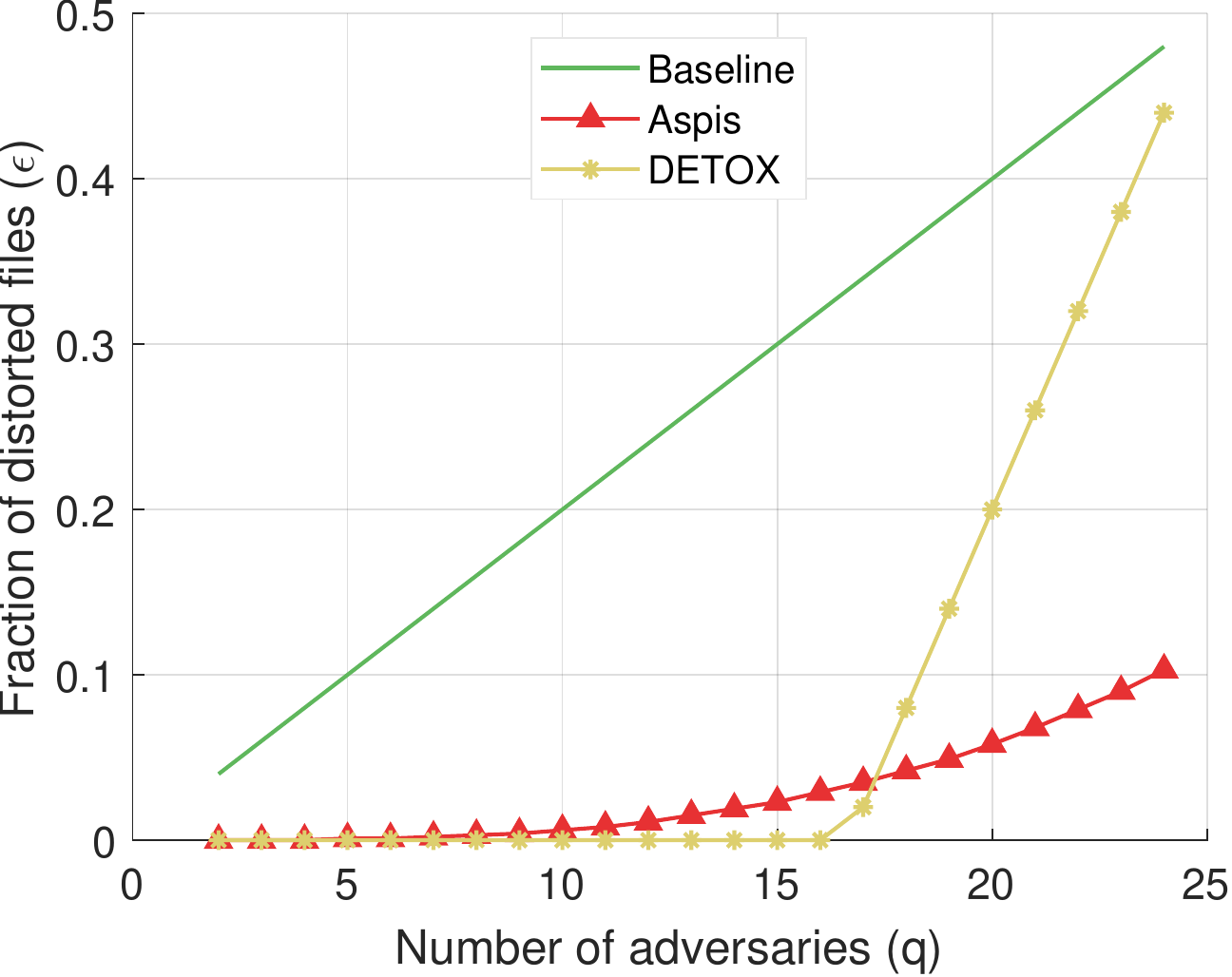}
		\caption{Weak attacks.}
		\label{fig:epsilon_simulations_weak_K50}
	\end{subfigure}
	\caption{Distortion fraction of optimal and weak attacks for $(K,r)=(50,3)$ and comparison.}
	\label{fig:epsilon_simulations_K50}
\end{figure}

\subsection{Optimal Adversarial Strategy}
\label{sec:optimal_subset_attack}
Our second scenario is strong and involves adversaries which collude in the ``best'' way possible while knowing the full details of our detection algorithm. In this discussion, we provide an upper bound on the number of files that can be corrupted in this case and demonstrate a strategy that the adversarial workers can follow to achieve this upper bound.


Let us index the $q$ adversaries in $A = \{A_1,A_2,\dots,A_q\}$ and the honest workers in $H$. We say that two workers $U_{j_1}$ and $U_{j_2}$ disagree if there is no edge between them in $\mathbf{G}$. The non-existence of an edge between $U_{j_1}$ and $U_{j_2}$ only means that they disagree in \emph{at least one} of the $\binom{K-2}{r-2}$ files that they jointly participate in. For corrupting the gradients, each adversary has to disagree on the computations with a subset of the honest workers. An adversary may also disagree with other adversaries.  Let $D_i$ denote the set of disagreement workers for adversary $A_i, i = 1,2,\dots, q$, where $D_i$ can contain members from $A$ and from $H$. 

Upon formation of $\mathbf{G}$ we know that a worker $U_j$ will be flagged as adversarial if $deg(U_j) < K - q -1$. Therefore to avoid detection, a \emph{necessary} condition is that $|D_j|\leq q$. We now upper bound the number of files that can be corrupted under any possible strategy employed by the adversaries. We fall back to robust aggregation in case of more than one maximum clique in $\mathbf{G}$. Then, a gradient can only be corrupted if a majority of the workers computing it are adversarial and agree on a wrong value. 

For a given file $F$, let $A' \subseteq A$ with $|A'| \geq r'$ be the set of ``active adversaries'' in it, i.e., $A' \subseteq F$ consists of Byzantines that collude to create a majority that distorts the gradient on it. In this case, the remaining workers in $F$ belong to $\cap_{i \in A'} D_i$, where we note that $|\cap_{i \in A'} D_i| \leq q$. Let $X_j, j = r',r'+1 \dots,r$ denote the subset of files where the set of active adversaries is of size $j$; note that $X_j$ depends on the disagreement sets $D_i, i = 1,2, \dots, q$. Formally,
\begin{eqnarray}
	X_j &=& \{F: \exists A' \subseteq A \cap F, |A'| = j,\nonumber\\
	&&\qquad\text{~and~} \forall~ U_j \in F \setminus A', U_j \in \cap_{i \in A'} D_i\}. \label{eq:X_j_files}\quad
\end{eqnarray}
Then, for a given choice of disagreement sets, the number of files that can be corrupted is given by $|\cup_{j=r'}^r X_j|$. We obtain an upper bound on the maximum number of corrupted files by maximizing this quantity with respect to the choice of $D_i, i = 1,2, \dots, q$, i.e.,
\begin{equation}
	c_{\mathrm{max}}^{(q)} = \max\limits_{D_i, |D_i| \leq q,  i = 1,2,\dots, q} |\cup_{j=r'}^r X_j|
\end{equation}
where the maximization is over the choices of the disagreement sets $D_1,D_2,\dots,D_q$. 
An intuitive adversarial strategy based on Eq. \eqref{eq:X_j_files} would be to maximize the set $\bigcap\limits_{i\in A' \cap F}D_i$ for every possible file $F$. In order to achieve this, for all groups $F$, the adversaries will need to fix a subset of $q$ non-adversaries, say $D \subset H$, which will be the set of workers with which all adversaries will disagree, i.e., $D_i = D$ for $i=1,2,\dots,q$. We present the following theorem (see Appendix Section \ref{appendix:fixed_diagreement_optimality} for proof).

\begin{theorem}
	\label{theorem:aspis_optimal_attack}
	Consider a training cluster of $K$ workers with $q$ adversaries using Algorithm \ref{alg:subset_placement} to assign the $f = \binom{K}{r}$ files to workers and Algorithm \ref{alg:detection} for adversary detection. Under an optimal adversary model the maximum number of files that can be corrupted is 
	\begin{equation}
		c_{\mathrm{max}}^{(q)} = \frac{1}{2}{2q\choose r}.
	\end{equation}
	Furthermore, if all adversaries fix a set $D \subset H$ of honest workers with which they will consistently disagree on the gradient (by distorting it), this upper bound can be achieved.
\end{theorem}
In particular, the proposed attack is optimal and there is no other attack that can corrupt more files under the Aspis algorithm. One such attack is carried out in Figure \ref{fig:subset_assignment_K7_r3_graph_failure} for the setup of Example \ref{ex:placement_K7_r3}. The adversaries $A = \{U_1, U_2, U_3\}$ consistently disagree with the workers in $D = \{U_4, U_5, U_6\} \subset H$. The ambiguity as to which of the two maximum cliques ($\{U_1, U_2, U_3, U_7\}$ or $\{U_4, U_5, U_6, U_7\}$) is the honest one makes an accurate detection impossible; robust aggregation will be performed instead.

\section{Distortion Fraction Evaluation}
\label{section:distortion_fraction_analysis}
We have performed simulations of the fraction of distorted files (defined as $\epsilon=c^{(q)}/f$) incurred by Aspis and other competing aggregators. The main motivation of this analysis is that our deep learning experiments (\emph{cf.} Section \ref{sec:experiments}) as well as prior work \cite{byzshield} show that $\epsilon$ serves as a surrogate of the model's convergence with respect to accuracy. In addition, our simulations show that Aspis enjoys values of $\epsilon$ which are as much as 99\% lower for the same $q$ compared to other techniques and this attests to our theoretical robustness guarantees. This comparison involves our work and state-of-the-art schemes under the best- and worst-case choice of the $q$ adversaries with respect to the achievable value of $\epsilon$. We compare our work with \emph{baseline} approaches which do not involve redundancy or majority voting. Their aggregation is applied directly to the $K$ gradients returned by the workers ($f=K$, $c_{\mathrm{max}}^{(q)}=q$ and $\epsilon=q/K$).

Let us first discuss the scenario of an \emph{optimal} attack. For Aspis, we used the proposed attack from Section \ref{sec:optimal_subset_attack} and the corresponding computation of $c^{(q),Aspis}$ of Theorem \ref{theorem:aspis_optimal_attack}. \emph{DETOX} in \cite{detox} employs a redundant assignment followed by majority voting and offers robustness guarantees which crucially rely on a ``random choice'' of the Byzantines. \cite{byzshield} have demonstrated the importance of a careful task assignment and observed that redundancy is not sufficient to yield Byzantine resilience dividends. They proposed an optimal choice of the $q$ Byzantines that maximizes $\epsilon^{DETOX}$ which we used in our experiments. In short, DETOX splits the $K$ workers into $K/r$ groups. All workers within a group process the same subset of the batch, specifically containing $br/K$ samples. This phase is followed by majority voting on a group-by-group basis. The authors of \cite{byzshield} suggested choosing the Byzantines such that at least $r'$ workers in each group are adversarial in order to distort the corresponding gradients. In this case, $c^{(q), DETOX} = \lfloor \frac{q}{r'} \rfloor$ and $\epsilon^{DETOX} = \lfloor \frac{q}{r'} \rfloor \times r/K$. We also compare with the distortion fraction incurred by ByzShield \cite{byzshield} under a worst-case scenario. For this scheme, there is no known optimal attack and its authors performed an exhaustive combinatorial search to find the $q$ adversaries that maximize $\epsilon^{ByzShield}$ among all possible options; we follow the same process here to simulate ByzShield's distortion fraction computation and utilize their scheme based on \emph{mutually orthogonal Latin squares}. The reader can refer to Figure \ref{fig:epsilon_simulations_optimal_K50} and Appendix Tables \ref{table:epsilon_simulations_K15}, \ref{table:epsilon_simulations_K21} and \ref{table:epsilon_simulations_K24} for our results. Aspis achieves major reductions in $\epsilon$; for instance, $\epsilon^{Aspis}$ is reduced by up to 99\% compared to both $\epsilon^{Baseline}$ and $\epsilon^{DETOX}$ in Figure \ref{fig:epsilon_simulations_optimal_K50}.

We will next introduce the utilized \emph{weak} attacks. For our scheme, we will make an arbitrary choice of $q$ adversaries which carry out the method introduced in Section \ref{sec:weak_subset_attack}, i.e., they will distort all files and a successful detection is possible. As discussed, the fraction of corrupted gradients is $\epsilon^{Aspis} = \binom{q}{r}/\binom{K}{r}$ in that case. For DETOX, a simple benign attack is used. To that end, let the $K/r$ files be $B_{t,0},B_{t,1},\dots,B_{t,K/r-1}$. Initialize $A=\emptyset$ and choose the $q$ Byzantines as follows: for $i=0,1,\dots,q-1$, among the remaining workers in $\{U_1,U_2,\dots,U_K\} - A$ add a worker from the group $B_{t,i \mod K/r}$ to the adversarial set $A$. Then,
\begin{equation*}
	c^{(q), DETOX} = \left\{
	\begin{array}{ll}
		q-\frac{K}{r}(r'-1) & \text{if } q > \frac{K}{r}(r'-1),\\
		0 & \text{otherwise}.
	\end{array}
	\right.
\end{equation*}
The results on this scenario are in Figure \ref{fig:epsilon_simulations_weak_K50}. 

For baseline schemes, there is no notion of ``weak'' or ``optimal'' attack with respect to the choice of the $q$ Byzantines; hence we can choose any subset of them achieving $\epsilon^{Baseline} = q/K$.


\begin{figure*}[t]
	\centering
	\begin{subfigure}[b]{0.33\textwidth}
		\centering
		\includegraphics[scale=0.4]{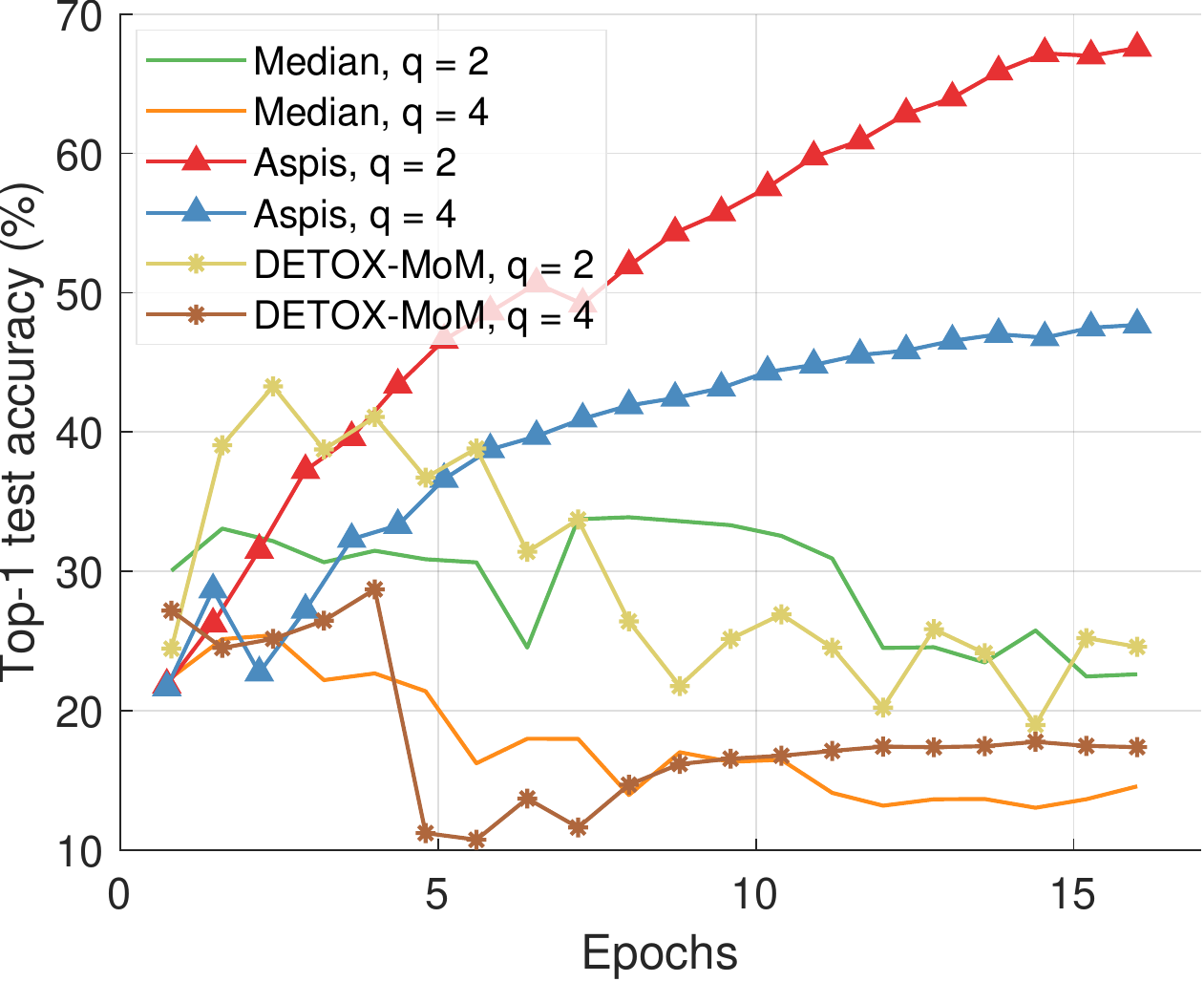}
		\caption{Median-based defenses.}
		\label{fig:top1_fig_94}
	\end{subfigure}
	\hfill
	\begin{subfigure}[b]{0.33\textwidth}
		\centering
		\includegraphics[scale=0.4]{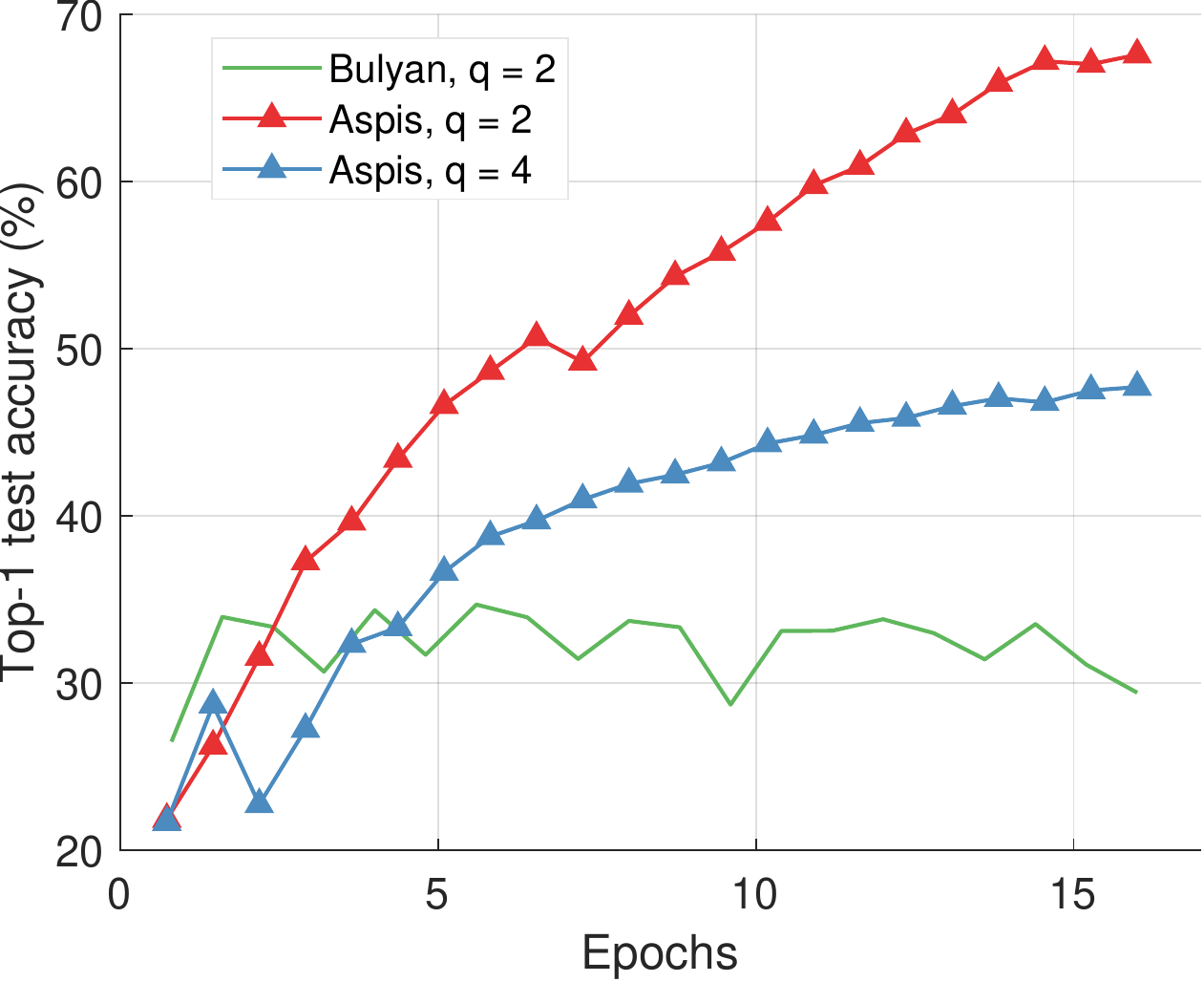}
		\caption{\emph{Bulyan}-based defenses.}
		\label{fig:top1_fig_95}
	\end{subfigure}
	\hfill
	\begin{subfigure}[b]{0.33\textwidth}
		\centering
		\includegraphics[scale=0.4]{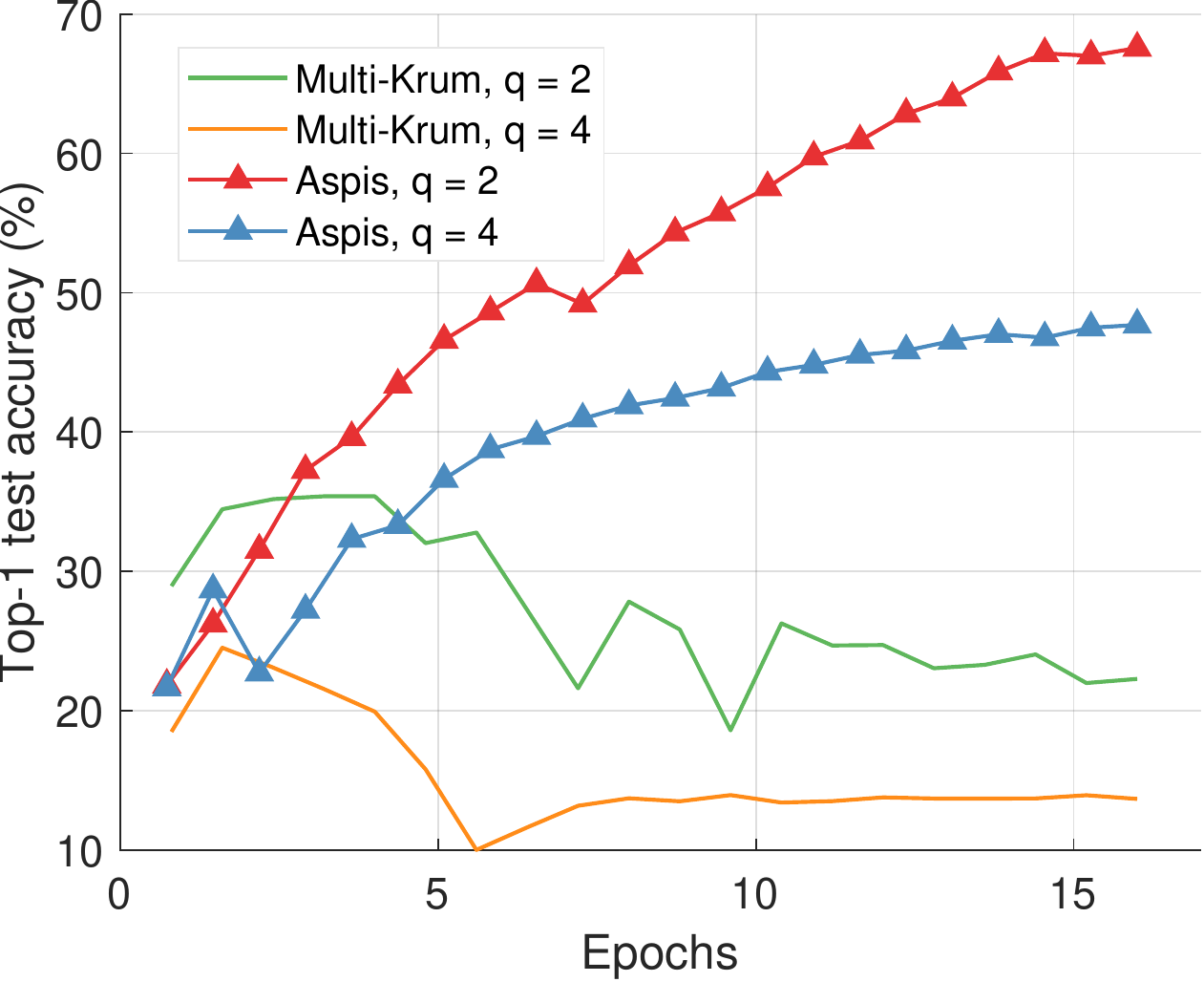}
		\caption{\emph{Multi-Krum}-based defenses.}
		\label{fig:top1_fig_96}
	\end{subfigure}
	\caption{\emph{ALIE} distortion under optimal attack scenarios (CIFAR-10), $K=15$.}
	\label{fig:top1_ALIE_optimal}
\end{figure*}

\begin{figure*}[t]
	\centering
	\begin{subfigure}[b]{0.33\textwidth}
		\centering
		\includegraphics[scale=0.4]{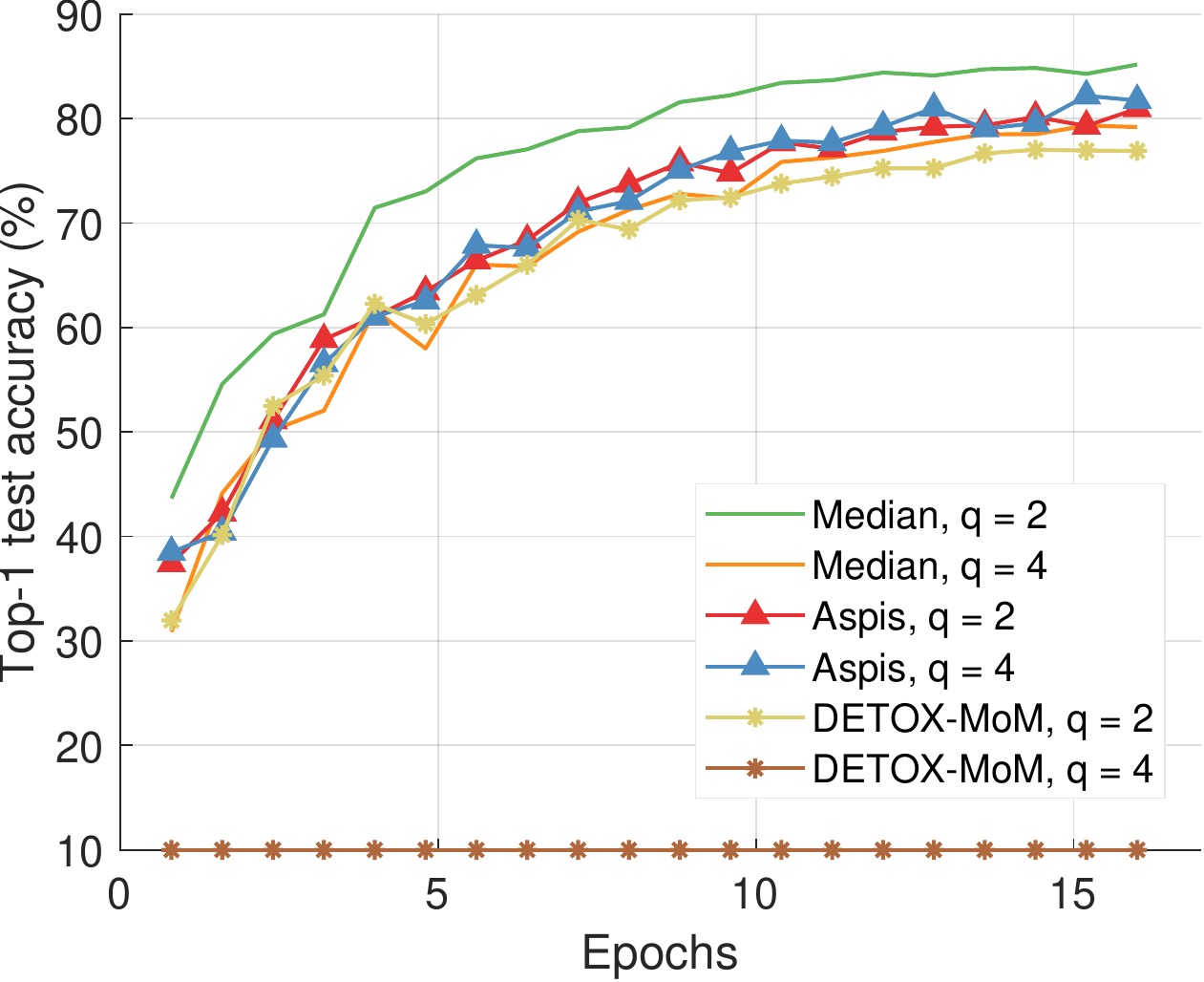}
		\caption{Median-based defenses.}
		\label{fig:top1_fig_97}
	\end{subfigure}
	\hfill
	\begin{subfigure}[b]{0.33\textwidth}
		\centering
		\includegraphics[scale=0.4]{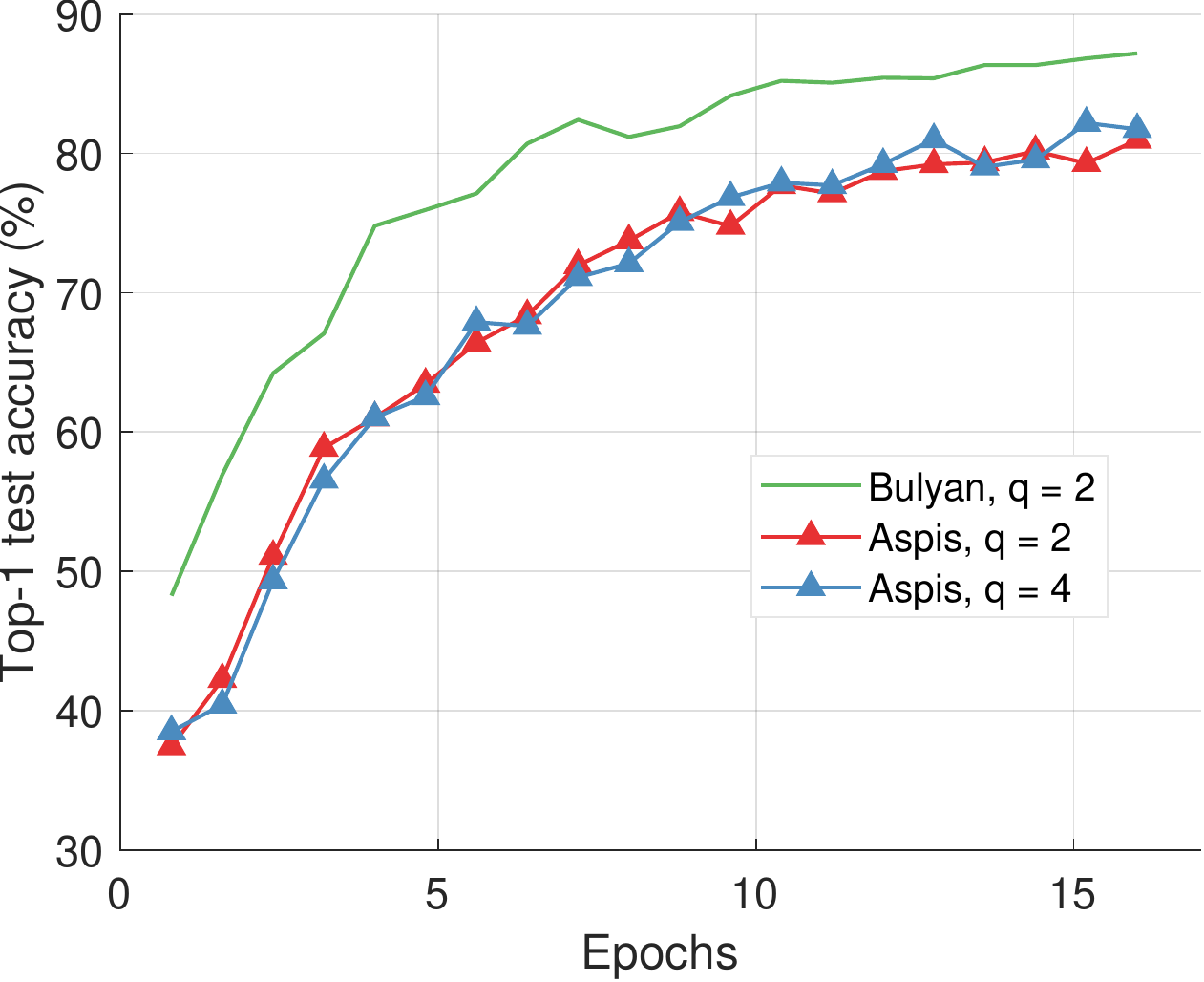}
		\caption{\emph{Bulyan}-based defenses.}
		\label{fig:top1_fig_98}
	\end{subfigure}
	\hfill
	\begin{subfigure}[b]{0.33\textwidth}
		\centering
		\includegraphics[scale=0.4]{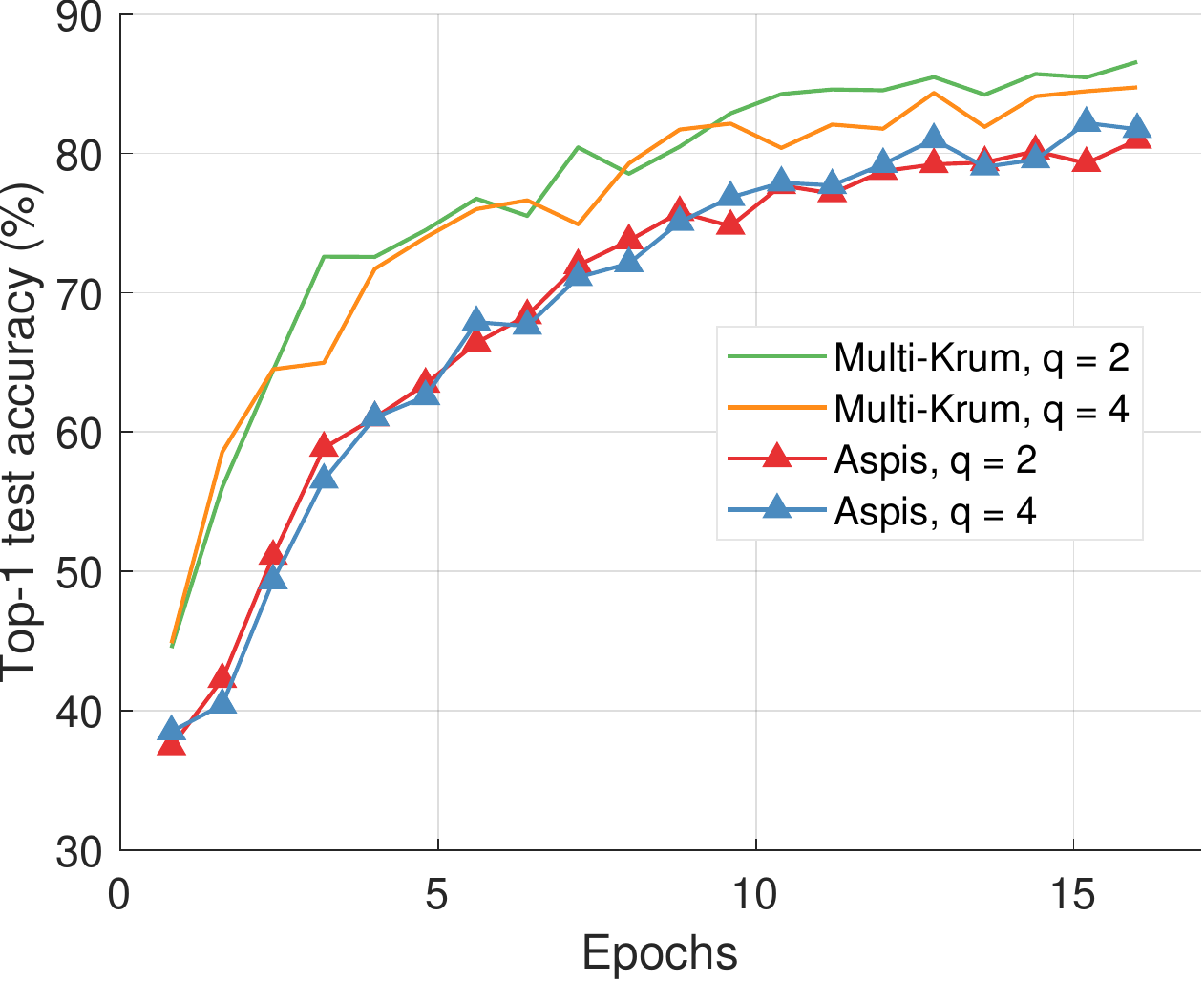}
		\caption{\emph{Multi-Krum}-based defenses.}
		\label{fig:top1_fig_99}
	\end{subfigure}
	\caption{\emph{Reversed gradient} distortion under optimal attack scenarios (CIFAR-10), $K=15$.}
\end{figure*}

\section{Large-scale Deep Learning Experiments}
\subsection{Experiment Setup}
\label{sec:experiment_setup}
We have evaluated the performance of our method and competing techniques in classification tasks on Amazon EC2 clusters. The project is written in PyTorch \cite{pytorch} and uses the MPICH library for communication between the different nodes. We worked with the CIFAR-10 dataset \cite{cifar10} using the ResNet-18 \cite{he_resnet} model. We used clusters of $K = 15$ and $21$ workers and redundancy $r = 3$. We simulated values of $q=2,4,6$ during training.
Detailed information about the implementation can be found in Appendix Section \ref{appendix:implementation_details}.

There are two different dimensions we experimentally evaluate with respect to the adversarial setup:\\
\textbf{1) Choice/orchestration of the adversaries}: This involves the different ways in which adversaries are chosen and can work together to inflict damage (\emph{cf.} weak and optimal attacks in Section \ref{sec:detection}).\\
\textbf{2) Gradient distortion methods}: This dimension is concerned with the method an adversary uses to distort the gradient value (\emph{cf.} Algorithm \ref{alg:main_algorithm}). We use a variety of state-of-the-art methods for distorting the computed gradients. 
\emph{ALIE} \cite{alie} involves communication among the Byzantines in which they jointly estimate the mean $\mu_i$ and standard deviation $\sigma_i$ of the batch's gradient for each dimension $i$ and subsequently use them to construct a distorted gradient that attempts to distort the median of the results. Another powerful attack is \emph{Fall of Empires (FoE)} \cite{FOE} which performs ``inner product manipulation'' to make the inner product between the true gradient and the robust estimator to be negative even when their distance is upper bounded by a small value. \emph{Reversed gradient} distortion returns $-c\*g$ for $c>0$, to the PS instead of the true gradient $\*g$. The ALIE algorithms is, to the best of our knowledge, the most sophisticated attack in literature. 

{\bf Competing methods:} We compare Aspis against the baseline implementations of median-of-means \cite{minsker2015}, Bulyan \cite{bulyan} and Multi-Krum \cite{blanchard_krum}. If $c_{\mathrm{max}}^{(q)}$ is the number of adversarial computations 
then Bulyan requires at least $4c_{\mathrm{max}}^{(q)}+3$ total number of computations while the same number for Multi-Krum is $2c_{\mathrm{max}}^{(q)}+3$. These constraints make these methods inapplicable for larger values of $q$ for which our method is robust. The second class of comparisons is with methods that use redundancy and specifically DETOX \cite{detox} for which we show that it can easily fail under malicious scenarios for large $q$. We compare with median-based techniques since they originate from robust statistics and are the base for many aggregators. DETOX is the most related redundancy-based work that is based on coding-theoretic techniques. Finally, Multi-Krum is a highly-cited aggregator that combines the intuitions of majority-based and squared-distance-based methods.


Note that for a baseline scheme all choices of $A$ are equivalent in terms of the value of $\epsilon$. In our comparisons with DETOX we will consider two attack scenarios with respect to the choice of the adversaries. For the \emph{optimal} choice in DETOX, we will use the method proposed in \cite{byzshield} and compare with the attack introduced in Section \ref{sec:optimal_subset_attack}. For the \emph{weak} one, we will choose the adversaries such that $\epsilon$ is minimized in DETOX and compare its performance with the scenario of Section \ref{sec:weak_subset_attack}.

\begin{figure*}[!htb]
	\minipage{0.32\textwidth}
	\includegraphics[width=0.9\linewidth]{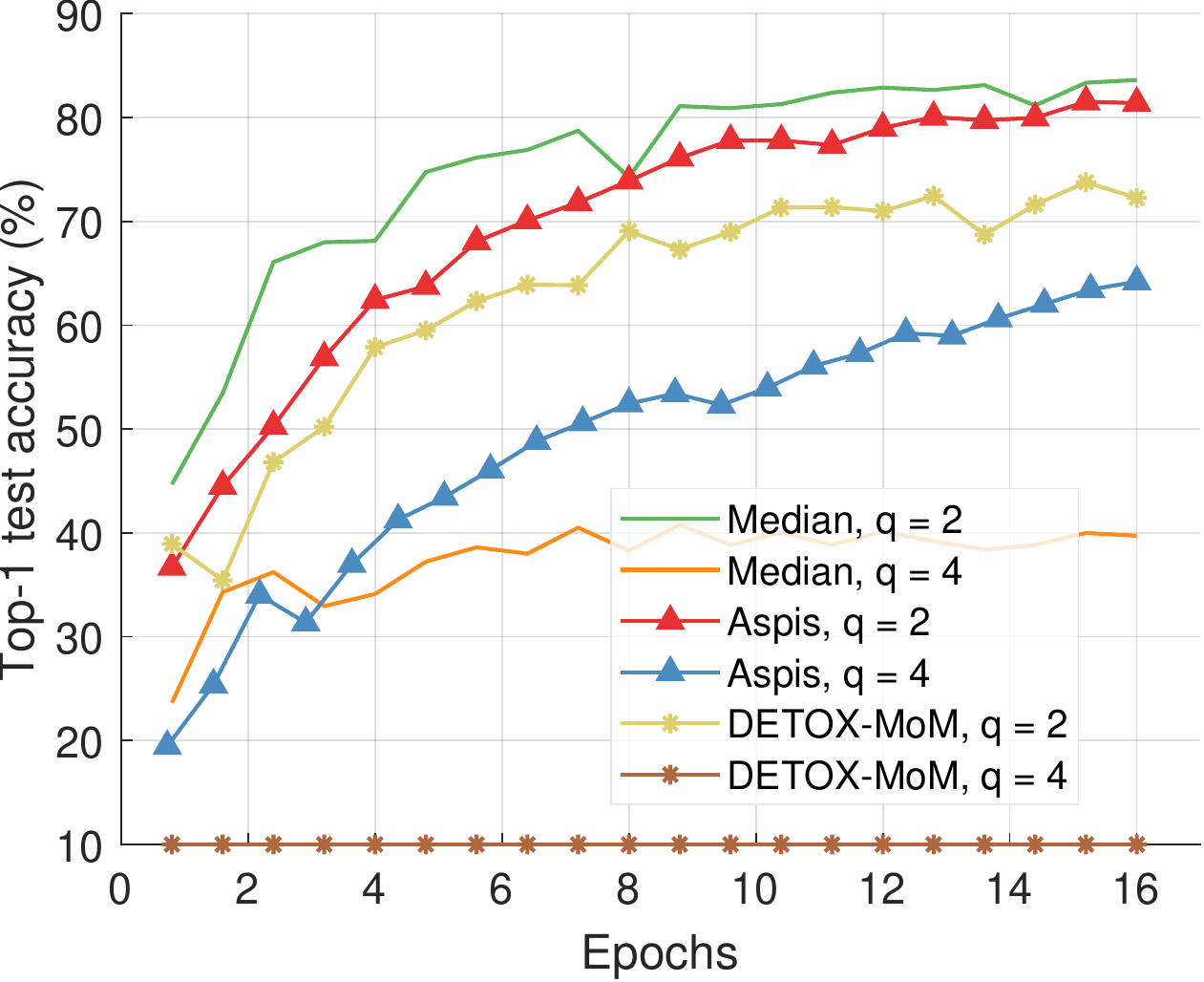}
	\caption{\emph{FoE} optimal attack and median-based defenses (CIFAR-10), $K=15$}
	\label{fig:top1_fig_101}
	\endminipage\hfill
	\minipage{0.32\textwidth}
	\includegraphics[width=0.9\linewidth]{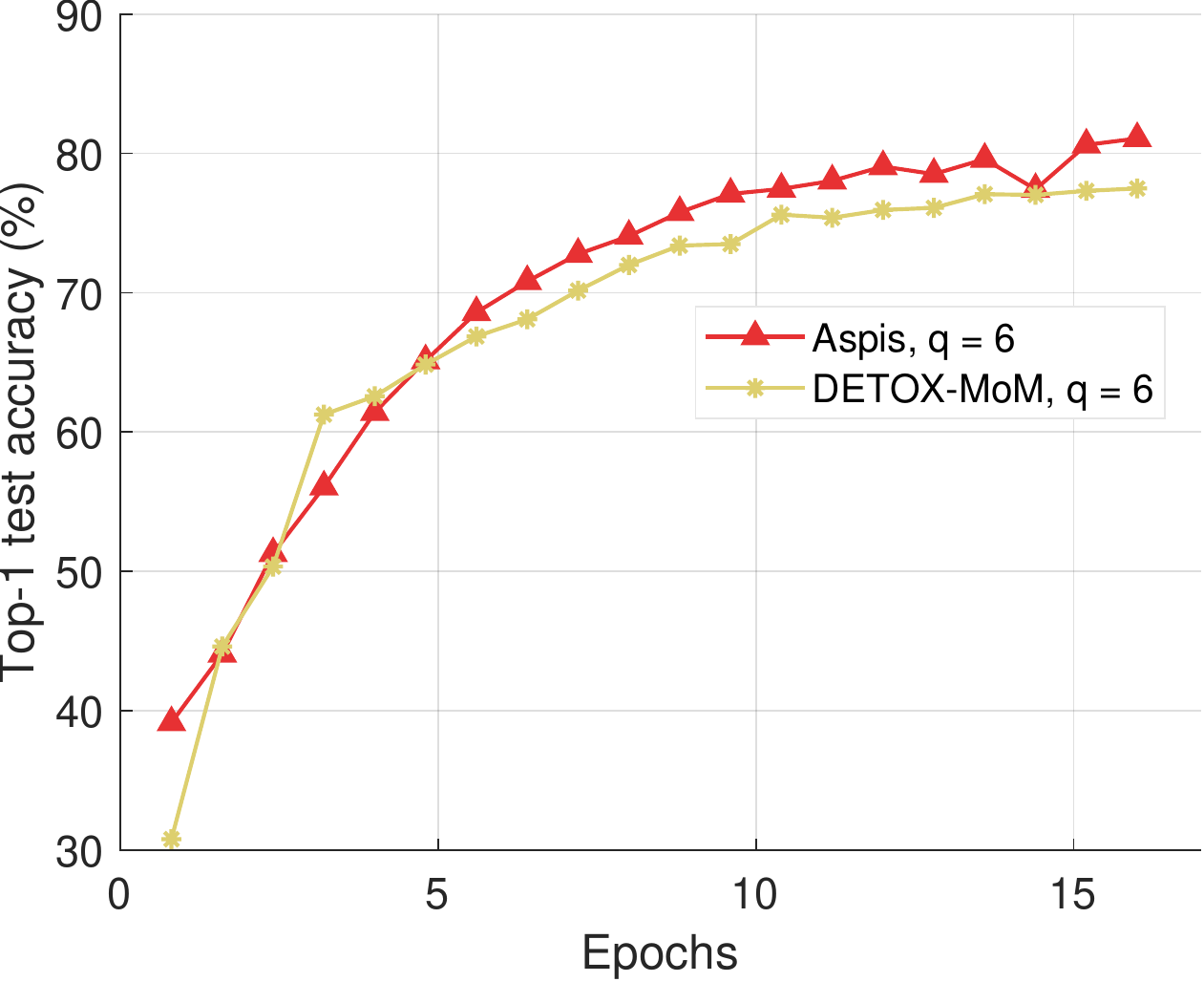}
	\caption{\emph{Reversed gradient} weak attack and median-based defenses (CIFAR-10), $K=15$.}
	\label{fig:top1_fig_100}
	\endminipage\hfill
	\minipage{0.32\textwidth}
	\includegraphics[width=0.9\linewidth]{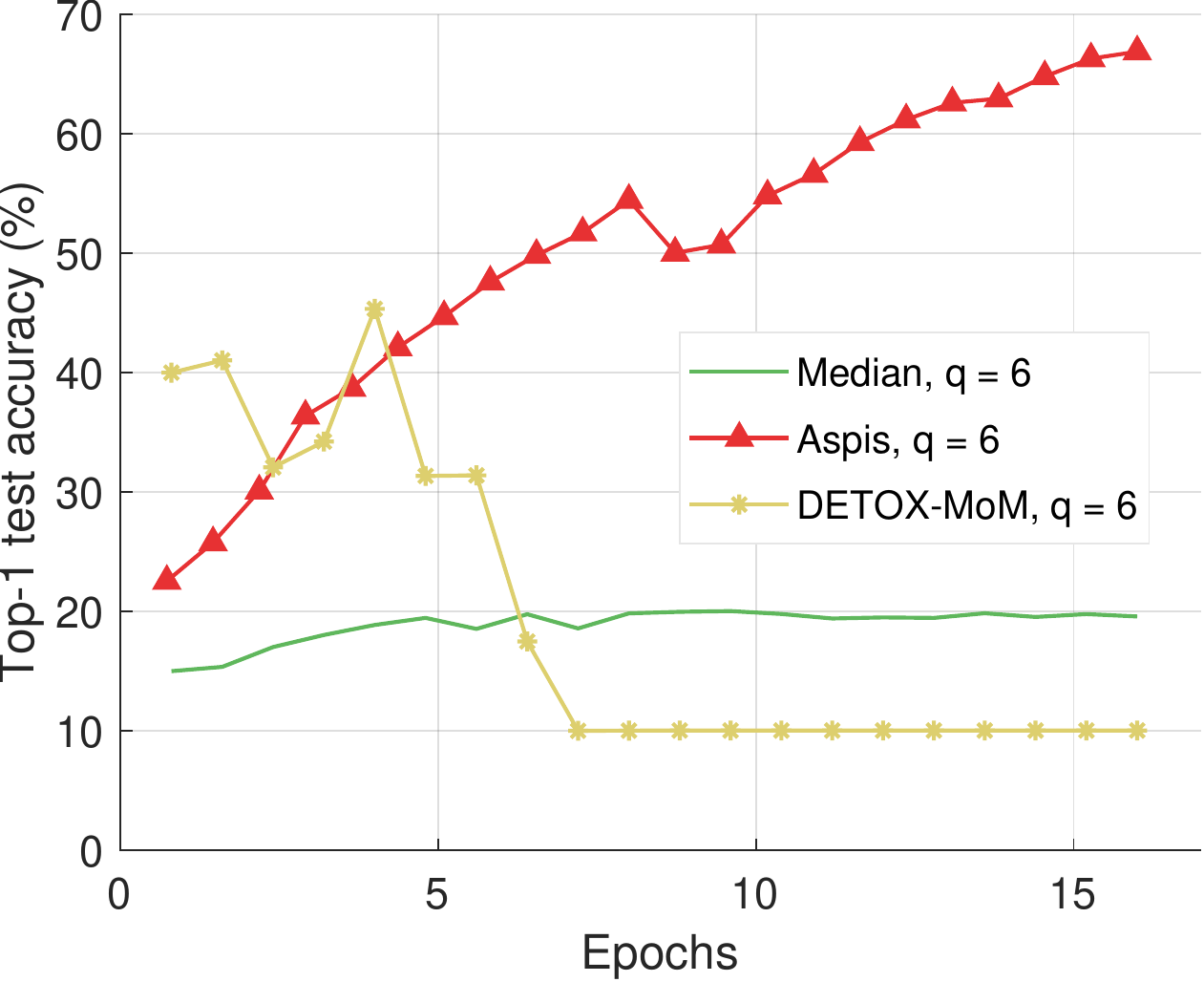}
	\caption{\emph{ALIE} weak attack and median-based defenses (CIFAR-10), $K=15$.}
	\label{fig:top1_fig_88}
	\endminipage
\end{figure*}

\begin{figure*}[t]
	\centering
	\begin{subfigure}[b]{0.33\textwidth}
		\centering
		\includegraphics[scale=0.4]{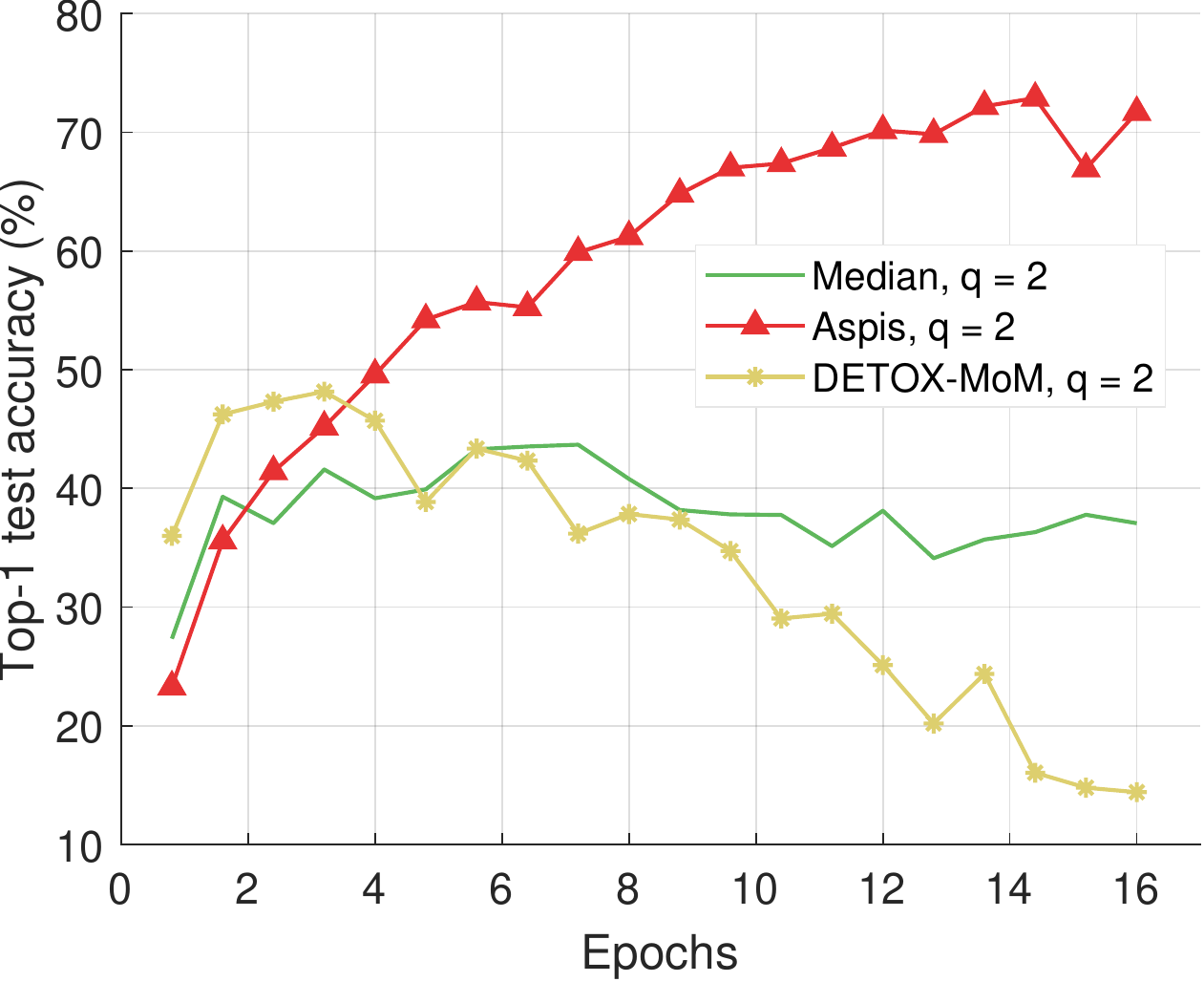}
		\caption{Median-based defenses.}
		\label{fig:top1_fig_102}
	\end{subfigure}
	\hfill
	\begin{subfigure}[b]{0.33\textwidth}
		\centering
		\includegraphics[scale=0.4]{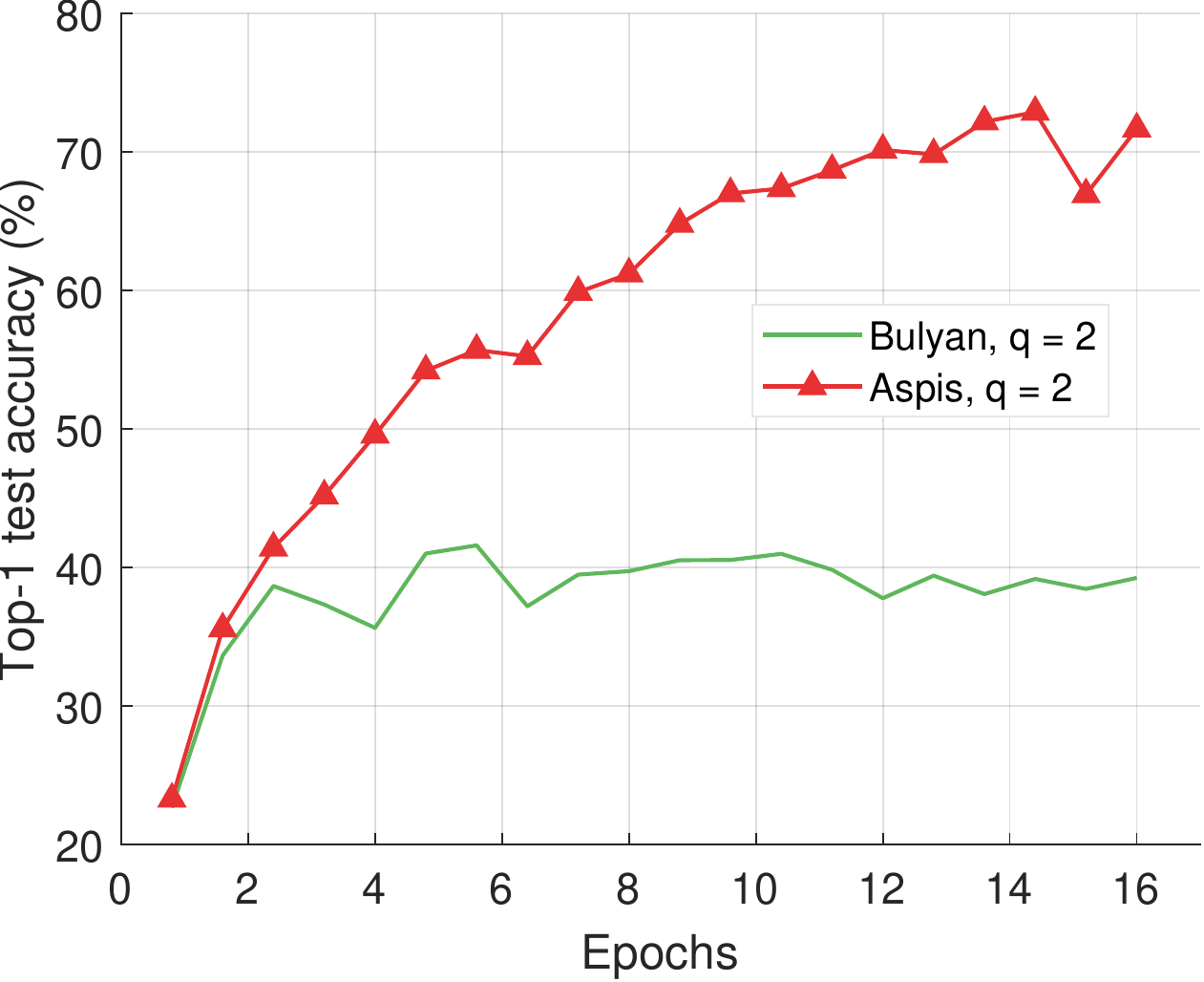}
		\caption{\emph{Bulyan}-based defenses.}
		\label{fig:top1_fig_103}
	\end{subfigure}
	\hfill
	\begin{subfigure}[b]{0.33\textwidth}
		\centering
		\includegraphics[scale=0.4]{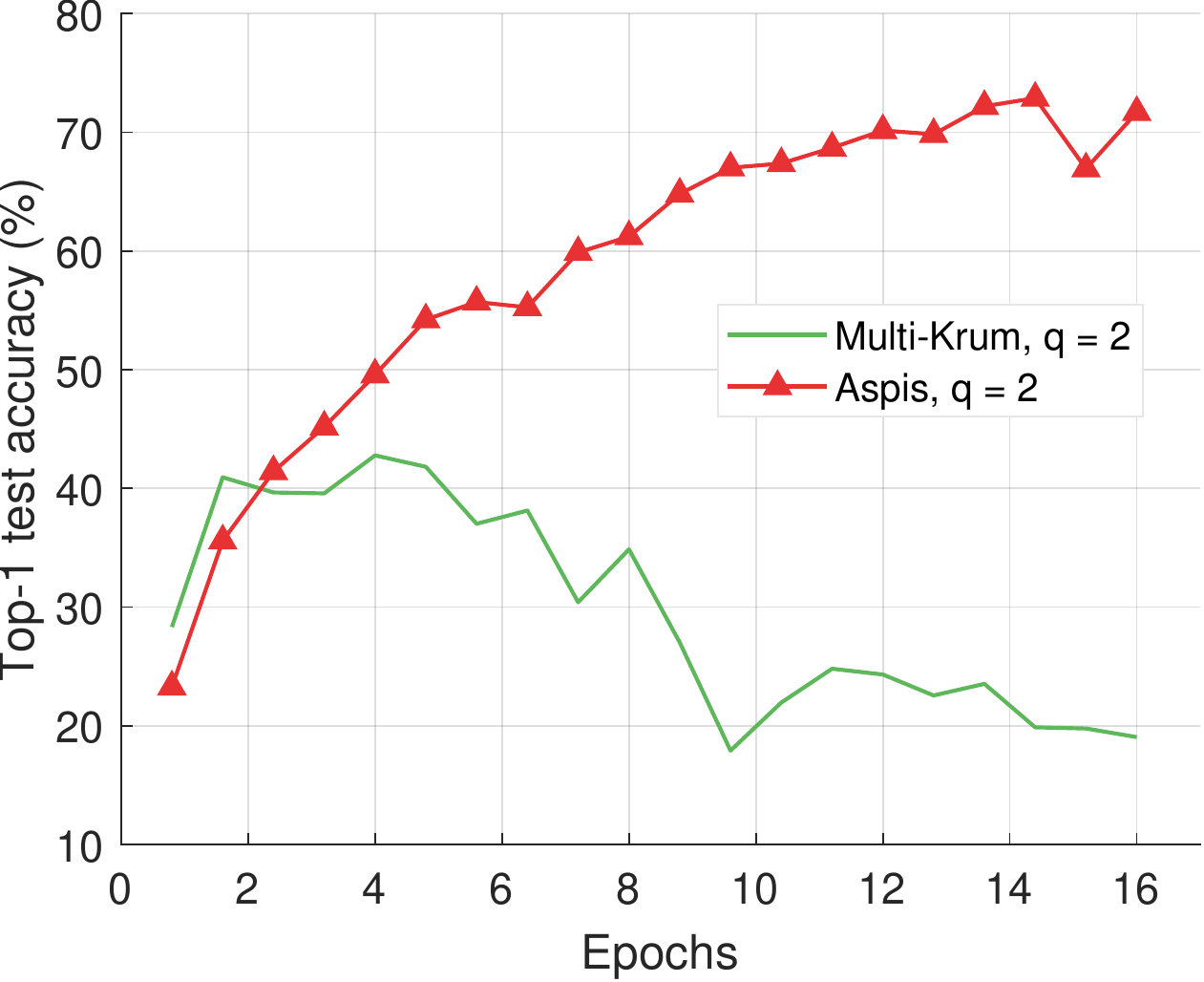}
		\caption{\emph{Multi-Krum}-based defenses.}
		\label{fig:top1_fig_104}
	\end{subfigure}
	\caption{\emph{ALIE} distortion under optimal attack scenarios (CIFAR-10), $K=21$.}
	\label{fig:top1_ALIE_optimal_K21}
\end{figure*}

\subsection{Experimental Results}
\label{sec:experiments}


\subsubsection{Comparison under Optimal Attacks}
We compare the different defense algorithms under optimal attack scenarios. Figure \ref{fig:top1_fig_94} compares our scheme Aspis with the baseline implementation of coordinate-wise median ($\epsilon=0.133, 0.267$ for $q=2,4$, respectively) and DETOX with median-of-means ($\epsilon=0.2, 0.4$ for $q=2,4$, respectively) under the ALIE attack. Aspis converges faster and achieves at least a 35\% average accuracy boost (at the end of the training) for both values of $q$ ($\epsilon^{Aspis}=0.004, 0.062$ for $q=2,4$, respectively).\footnote{Please refer to Appendix Tables \ref{table:epsilon_simulations_K15} and \ref{table:epsilon_simulations_K21} for the values of the distortion fraction $\epsilon$ each scheme incurs.} In Figures \ref{fig:top1_fig_95} and \ref{fig:top1_fig_96} we observe similar trends in our experiments with Bulyan and Multi-Krum where Aspis significantly outperforms these techniques. For the current setup, Bulyan is not applicable for $q=4$ since $K = 15 < 4c_{\mathrm{max}}^{(q)}+3 = 4q+3 = 19$. Also, neither Bulyan nor Multi-Krum can be paired with DETOX for $q \geq 1$ since the inequalities $f \geq 4c_{\mathrm{max}}^{(q)}+3$ and $f \geq 2c_{\mathrm{max}}^{(q)}+3$ cannot be satisfied. Please refer to Section \ref{sec:experiment_setup} and Section \ref{section:distortion_fraction_analysis} for more details on these requirements. Also, note that the accuracy of most competing methods fluctuates more than in the results presented in the corresponding papers \cite{detox} and \cite{alie}. This is expected as we consider stronger attacks compared to those papers, i.e., optimal deterministic attack on DETOX and, in general, up to 27\% adversarial workers in the cluster. Also, we have done multiple experiments with different random seeds to demonstrate the stability and superiority of our accuracy results compared to other methods (against median-based defenses in Appendix Figure \ref{fig:top1_error_bar_median}, Bulyan in Figure \ref{fig:top1_error_bar_bulyan} and Multi-Krum in Figure \ref{fig:top1_error_bar_multikrum}); we point the reader to Appendix Section \ref{appendix:error_bars} for this analysis. This analysis is clearly missing from most prior work, including that of ALIE \cite{alie}, and their presented results are only a snapshot of a single experiment which may or may not be reproducible as is. The results for the reversed gradient attack are shown in Figures \ref{fig:top1_fig_97}, \ref{fig:top1_fig_98} and \ref{fig:top1_fig_99}. Given that this is a much weaker attack \cite{byzshield,detox} all of the schemes considered, including the baseline methods, are expected to perform well; indeed in most of the cases, the model converges to approximately 80\% accuracy. However, DETOX fails to converge to high accuracy for $q=4$ as in the case of ALIE; one explanation is that $\epsilon^{DETOX}=0.4$ for $q=4$. Under the Fall of Empires (FoE) distortion (\emph{cf.} Figure \ref{fig:top1_fig_101}) our method still enjoys an accuracy advantage over the baseline and DETOX schemes which becomes more important as the number of Byzantines in the cluster increases.

\subsubsection{Comparison under Weak Attacks}
For baseline schemes, the discussion of weak versus optimal choice of the adversaries is not very relevant as any choice of the $q$ Byzantines can overall distort at most $q$ out of the $K$ gradients. Hence, for weak scenarios, we chose to compare mostly with DETOX. The accuracy is reported on Figures \ref{fig:top1_fig_100} and \ref{fig:top1_fig_88} according to which Aspis shows an improvement under attacks on the more challenging end of the spectrum (ALIE). According to Appendix Table \ref{table:epsilon_simulations_weak_K15}, Aspis enjoys a fraction $\epsilon^{Aspis} = 0.044$ while $\epsilon^{Baseline} = 0.4$ and $\epsilon^{DETOX} = 0.2$ for $q=6$.

Our experiments on a larger cluster of $K=21$ workers under the ALIE attack can be found in Figure \ref{fig:top1_ALIE_optimal_K21}.

\section{Conclusions and Future Work}
In this work we have presented Aspis, a Byzantine-resilient distributed scheme that uses redundancy and robust aggregation in novel ways to detect adversarial behavior by the workers. Our theoretical analysis and numerical experiments clearly indicate the superior performance of Aspis compared to state-of-the-art. Our experiments show that Aspis requires increased computation and communication time as compared to prior work, e.g., note that each worker has to transmit $l$ gradients instead of $1$ in related work \cite{detox,draco} (see Appendix Section \ref{appendix:computation_communication_overhead} for details). We emphasize however, that under strong attacks (optimal Byzantine behavior) Aspis converges to high accuracy while competing methods will never converge irrespective of how long the algorithm runs for (please refer to Figures \ref{fig:top1_ALIE_optimal}, \ref{fig:top1_fig_97} and \ref{fig:top1_fig_101}); for a strong distortion attack like ALIE, this is evident even under weak adversarial strategies with respect to the Byzantine node behavior (cf. Figure \ref{fig:top1_fig_88}). 

Our experiments involve clusters of up to $21$ workers. As we scale our solution to more workers, the total number of files and the computation load $l$ of each worker will also scale; this increases the memory needed to store the gradients during aggregation. For complex neural networks, the memory to store the model and the intermediate gradient computations is by far the most memory-consuming aspect of the algorithm. For these reasons, our scheme is mostly suitable to training large datasets using fairly compact models that do not require too much memory. Also, there are opportunities for reducing the time overhead. For instance, utilizing GPUs and communication-related algorithmic improvements are worth exploring. Finally, convergence analysis is under investigation and space limitations do not allow us to discuss it here.




\clearpage

\bibliographystyle{mlsys2022}



\clearpage
\appendix
\section{Appendix}
\subsection{Asymptotic Complexity}
\label{appendix:asymptotic}
If the gradient computation has linear complexity (assuming $\mathcal{O}(1)$ cost for the gradient computation with respect to one model parameter) and since each worker is assigned to $l$ files of $b/f$ samples each, the gradient computation cost at the worker level is $\mathcal{O}((lb/f)d)$ ($K$ such computations in parallel). In our schemes, however, $b$ is a constant multiple of $f$ and in general $r < l = \binom{K-1}{r-1}$; hence, the complexity becomes $\mathcal{O}(ld)$ which is similar to other redundancy-based schemes in \cite{byzshield,detox,draco}. The clique-finding problem that follows as part of our detection is NP-complete. However, our experimental evidence suggests that for the kind of graphs we construct this computation takes a infinitesimal fraction of the execution time. The NetworkX package \cite{networkx} which we use for enumerating all maximal cliques is based on the algorithm of \cite{tomita_clique} and has asymptotic complexity $\mathcal{O}(3^{K/3})$. We provide extensive simulations of the clique enumeration time under the Aspis file assignment for $K = 100$ and redundancy $r=5$ (\emph{cf.} Tables \ref{table:clique_time_weak}, \ref{table:clique_time_optimal} for weak and optimal attack as introduced in Section \ref{sec:detection}, respectively). We emphasize that this value of $K$ exceeds by far the typical values of $K$ of prior work and the number of servers would suffice for the majority of challenging training tasks. Even in this case, the cost of enumerating all cliques is negligible. For this experiment, we used an EC2 instance of type \texttt{i3.16xlarge}. The complexity of robust aggregation varies significantly depending on the operator. For example, majority voting can be done in time which scales linearly with the number of votes using \emph{MJRTY} proposed in \cite{boyer_majority}. In our case, this is $\mathcal{O}(Kd)$ as the PS needs to use the $d$-dimensional input from all $K$ machines. Krum \cite{blanchard_krum}, Multi-Krum \cite{blanchard_krum} and Bulyan \cite{bulyan} are applied to all $K$ workers by default and require $\mathcal{O}(K^2(d+\mathrm{log}K))$.

\begin{table*}[!t]
	\large
	\newcommand\Ksubtablewidth{0.43\linewidth}
	\newcommand\Kresizetabular{\columnwidth}
	\centering
	\captionsetup[subtable]{position = below}
	\caption{Distortion fraction of optimal and weak attacks for $(K,f,l,r)=(15,455,91,3)$ and comparison.}
	\label{table:epsilon_simulations_K15}
	\begin{subtable}{\Ksubtablewidth}
		\centering
		{
			\resizebox{0.95\Kresizetabular}{!}{
				\begin{tabular}{P{0.5cm} R R R R}
					\toprule
					\multicolumn{1}{c}{$q$} & \multicolumn{1}{c}{$\epsilon^{Aspis}$} & \multicolumn{1}{c}{$\epsilon^{Baseline}$} & \multicolumn{1}{c}{$\epsilon^{DETOX}$} & \multicolumn{1}{c}{$\epsilon^{ByzShield}$} \\
					\hline
					$2$ & 0.004 & 0.133 & 0.2 & 0.04 \\
					$3$ & 0.022 & 0.2 & 0.2 & 0.12 \\
					$4$ & 0.062 & 0.267 & 0.4 & 0.2 \\
					$5$ & 0.132 & 0.333 & 0.4 & 0.32 \\
					$6$ & 0.242 & 0.4 & 0.6 & 0.48 \\
					$7$ & 0.4 & 0.467 & 0.6 & 0.56 \\
					\toprule
				\end{tabular}
			}
			\caption{Optimal attacks.}
		}
	\end{subtable}
	\hspace{0.01\textwidth}
	\begin{subtable}{\Ksubtablewidth}
		\centering
		{
			\resizebox{0.72\Kresizetabular}{!}{
				\begin{tabular}{P{0.5cm} R R R}
					\toprule
					$q$ & \multicolumn{1}{c}{$\epsilon^{Aspis}$} & \multicolumn{1}{c}{$\epsilon^{Baseline}$} & \multicolumn{1}{c}{$\epsilon^{DETOX}$} \\
					\hline
					$2$ & 0.002 & 0.133 & 0 \\
					$3$ & 0.002 & 0.2 & 0 \\
					$4$ & 0.009 & 0.267 & 0 \\
					$5$ & 0.022 & 0.333 & 0 \\
					$6$ & 0.044 & 0.4 & 0.2 \\
					$7$ & 0.077 & 0.467 & 0.4 \\
					\toprule
				\end{tabular}
			}
			\caption{Weak attacks.}
			\label{table:epsilon_simulations_weak_K15}
		}
	\end{subtable}
\end{table*}

\begin{table*}[!t]
	\large
	\newcommand\Ksubtablewidth{0.43\linewidth}
	\newcommand\Kresizetabular{\columnwidth}
	\centering
	\captionsetup[subtable]{position = below}
	\caption{Distortion fraction of optimal and weak attacks for $(K,f,l,r)=(21,1330,190,3)$ and comparison.}
	\label{table:epsilon_simulations_K21}
	\begin{subtable}{\Ksubtablewidth}
		\centering
		{
			\resizebox{0.95\Kresizetabular}{!}{
				\begin{tabular}{P{0.6cm} R R R R}
					\toprule
					$q$ & \multicolumn{1}{c}{$\epsilon^{Aspis}$} & \multicolumn{1}{c}{$\epsilon^{Baseline}$} & \multicolumn{1}{c}{$\epsilon^{DETOX}$} & \multicolumn{1}{c}{$\epsilon^{ByzShield}$} \\
					\hline
					$2$ & 0.002 & 0.095 & 0.143 & 0.02 \\
					$3$ & 0.008 & 0.143 & 0.143 & 0.06 \\
					$4$ & 0.021 & 0.19 & 0.286 & 0.1 \\
					$5$ & 0.045 & 0.238 & 0.286 & 0.16 \\
					$6$ & 0.083 & 0.286 & 0.429 & 0.24 \\
					$7$ & 0.137 & 0.333 & 0.429 & 0.33 \\
					$8$ & 0.211 & 0.381 & 0.571 & 0.43 \\
					$9$ & 0.307 & 0.429 & 0.571 & 0.51 \\
					$10$ & 0.429 & 0.476 & 0.714 & 0.59 \\
					\toprule
				\end{tabular}
			}
			\caption{Optimal attacks.}
		}
	\end{subtable}
	\hspace{0.01\textwidth}
	\begin{subtable}{\Ksubtablewidth}
		\centering
		{
			\resizebox{0.72\Kresizetabular}{!}{
				\begin{tabular}{P{0.6cm} R R R}
					\toprule
					$q$ & \multicolumn{1}{c}{$\epsilon^{Aspis}$} & \multicolumn{1}{c}{$\epsilon^{Baseline}$} & \multicolumn{1}{c}{$\epsilon^{DETOX}$} \\
					\hline
					$2$ & 0.001 & 0.095 & 0 \\
					$3$ & 0.001 & 0.143 & 0 \\
					$4$ & 0.003 & 0.19 & 0 \\
					$5$ & 0.008 & 0.238 & 0 \\
					$6$ & 0.015 & 0.286 & 0 \\
					$7$ & 0.026 & 0.333 & 0 \\
					$8$ & 0.042 & 0.381 & 0.143 \\
					$9$ & 0.063 & 0.429 & 0.286 \\
					$10$ & 0.09 & 0.476 & 0.429 \\
					\toprule
				\end{tabular}
			}
			\caption{Weak attacks.}
		}
	\end{subtable}
\end{table*}

\begin{table*}[!t]
	\large
	\newcommand\Ksubtablewidth{0.43\linewidth}
	\newcommand\Kresizetabular{\columnwidth}
	\centering
	\captionsetup[subtable]{position = below}
	\caption{Distortion fraction of optimal and weak attacks for $(K,f,l,r)=(24,2024,253,3)$ and comparison.}
	\label{table:epsilon_simulations_K24}
	\begin{subtable}{\Ksubtablewidth}
		\centering
		{
			\resizebox{0.95\Kresizetabular}{!}{
				\begin{tabular}{P{0.6cm} R R R R}
					\toprule
					$q$ & \multicolumn{1}{c}{$\epsilon^{Aspis}$} & \multicolumn{1}{c}{$\epsilon^{Baseline}$} & \multicolumn{1}{c}{$\epsilon^{DETOX}$} & \multicolumn{1}{c}{$\epsilon^{ByzShield}$} \\
					\hline
					$2$ & 0.001 & 0.083 & 0.125 & 0.031 \\
					$3$ & 0.005 & 0.125 & 0.125 & 0.063 \\
					$4$ & 0.014 & 0.167 & 0.25 & 0.125 \\
					$5$ & 0.03 & 0.208 & 0.25 & 0.188 \\
					$6$ & 0.054 & 0.25 & 0.375 & 0.281 \\
					$7$ & 0.09 & 0.292 & 0.375 & 0.375 \\
					$8$ & 0.138 & 0.333 & 0.5 & 0.5 \\
					$9$ & 0.202 & 0.375 & 0.5 & 0.5 \\
					$10$ & 0.282 & 0.417 & 0.625 & 0.531 \\
					$11$ & 0.38 & 0.458 & 0.625 & 0.625\\
					\toprule
				\end{tabular}
			}
			\caption{Optimal attacks.}
		}
	\end{subtable}
	\hspace{0.01\textwidth}
	\begin{subtable}{\Ksubtablewidth}
		\centering
		{
			\resizebox{0.72\Kresizetabular}{!}{
				\begin{tabular}{P{0.6cm} R R R}
					\toprule
					$q$ & \multicolumn{1}{c}{$\epsilon^{Aspis}$} & \multicolumn{1}{c}{$\epsilon^{Baseline}$} & \multicolumn{1}{c}{$\epsilon^{DETOX}$} \\
					\hline
					$2$ & 0 & 0.083 & 0 \\
					$3$ & 0 & 0.125 & 0 \\
					$4$ & 0.002 & 0.167 & 0 \\
					$5$ & 0.005 & 0.208 & 0 \\
					$6$ & 0.01 & 0.25 & 0 \\
					$7$ & 0.017 & 0.292 & 0 \\
					$8$ & 0.028 & 0.333 & 0 \\
					$9$ & 0.042 & 0.375 & 0.125 \\
					$10$ & 0.059 & 0.417 & 0.25 \\
					$11$ & 0.082 & 0.458 & 0.375 \\
					\toprule
				\end{tabular}
			}
			\caption{Weak attacks.}
		}
	\end{subtable}
\end{table*}

\begin{table*}[!h]
	\large
	\newcommand\Ksubtablewidth{0.35\linewidth}
	\newcommand\Kresizetabular{\columnwidth}
	\centering
	\captionsetup[subtable]{position = below}
	\caption{Clique enumeration time in Aspis graph of $K=100$ vertices and redundancy $r=5$.}	
	\begin{subtable}{\Ksubtablewidth}
		\centering
		{
			\resizebox{0.95\Kresizetabular}{!}{
				\begin{tabular}{P{0.9cm}P{5cm}}
					\hline
					$q$ & Time (milliseconds) \\
					\hline
					5 & 9\\
					\hline
					15 & 7\\
					\hline
					25 & 5\\
					\hline
					35 & 5\\
					\hline
					45 & 5\\
					\hline
				\end{tabular}
			}
			\caption{Adversaries carry out weak attack.}
			\label{table:clique_time_weak}
		}
	\end{subtable}
	\hspace{0.07\textwidth}
	\begin{subtable}{\Ksubtablewidth}
		\centering
		{
			\resizebox{0.95\Kresizetabular}{!}{
				\begin{tabular}{P{0.9cm}P{5cm}}
					\hline
					$q$ & Time (milliseconds) \\
					\hline
					5 & 11\\
					\hline
					15 & 11\\
					\hline
					25 & 9\\
					\hline
					35 & 8\\
					\hline
					45 & 6\\
					\hline
				\end{tabular}
			}
			\caption{Adversaries carry out optimal attack.}
			\label{table:clique_time_optimal}
		}
	\end{subtable}
\end{table*}

\subsection{Proof of Theorem \ref{theorem:aspis_optimal_attack}}
\label{appendix:fixed_diagreement_optimality}
With $X_j$ given in (\ref{eq:X_j_files}), assuming $q\geq r'$, the number of distorted files is upper bounded by 
\begin{align}
	|\cup_{j=r'}^r X_j| &\leq \sum_{j=r'}^r |X_j| \text{~(by the union bound).} \label{eq:union_bd}
\end{align}
For that, recall that $r'=r(r+1)/2$ and that an adversarial majority of at least $r'$ distorted computations for a file is needed to corrupt that particular file. Note that $X_j$ consists of those files where the active adversaries $A'$ are of size $j$; these can be chosen in $\binom{q}{j}$ ways. The remaining workers in the file belong to $\cap_{i \in A'} D_i$ where $|\cap_{i \in A'} D_i| \leq q$. Thus, the remaining workers can be chosen in at most $\binom{q}{r-j}$ ways. It follows that
\begin{align}
	|X_j| \leq \binom{q}{j}\binom{q}{r-j}. \label{eq:upper_bd_X_j}
\end{align}

Therefore,

\begin{eqnarray}
	c_{\mathrm{max}}^{(q)} &\leq& {q \choose r'}{q \choose r-r'} + {q \choose r'+1}{q \choose r-(r'+1)}\nonumber\\
	&&+ \cdots \nonumber\\
	&&+ {q \choose r-1}{q \choose r-(r-1)} + {q \choose r}\label{eq:c_q_max_first_inequality}\\
	&=& \sum_{i=r'}^q{{q}\choose{i}}{{q}\choose{r-i}} \label{eq:optimal_c_q_max}\\
	&=& \sum_{i=0}^q{{q}\choose{i}}{{q}\choose{r-i}} - \sum_{i=0}^{r'-1}{{q}\choose{i}}{{q}\choose{r-i}}\label{eq:optimal_c_q_max_simplification_1}\\
	&=& \frac{1}{2}{2q\choose r}\label{eq:optimal_c_q_max_simplification_2}.
\end{eqnarray}
Eq. \eqref{eq:optimal_c_q_max} follows from the convention that ${n\choose k} = 0$ when $k > n$ or $k < 0$. Eq. \eqref{eq:optimal_c_q_max_simplification_2} follows from Eq. \eqref{eq:optimal_c_q_max_simplification_1} using the following observations

\begin{itemize}
	\item $\sum_{i=0}^q{{q}\choose{i}}{{q}\choose{r-i}} = \sum_{i=0}^r{{q}\choose{i}}{{q}\choose{r-i}} = {2q\choose r}$ in which the first equality is straightforward to show by taking all possible cases: $q<r$, $q=r$ and $q>r$.
	\item By symmetry, $\sum_{i=0}^{r'-1}{{q}\choose{i}}{{q}\choose{r-i}} = \sum_{i=r'}^{q}{{q}\choose{i}}{{q}\choose{r-i}} = \frac{1}{2}{2q\choose r}$.
\end{itemize}

The upper bound in Eq. \eqref{eq:c_q_max_first_inequality} is met with equality when all adversaries choose the same disagreement set which is a $q$-sized subset of the honest workers, i.e., $D_i = D \subset H$ for $i =1, \dots,q$. In this case, it can be seen that the sets $X_j, j=r', \dots, r$ are disjoint, so that \eqref{eq:union_bd} is met with equality. Moreover, \eqref{eq:upper_bd_X_j} is also an equality. This finally implies that \eqref{eq:c_q_max_first_inequality} is also an equality, i.e., this choice of disagreement sets saturates the upper bound.

It can also be seen that in this case the adversarial strategy yields a graph $\mathbf{G}$ with multiple maximum cliques. 
To see this, we note that the adversaries in $A$ agree with all the computed gradients in $H \setminus D$. Thus, they form of a clique of $M_{\mathbf{G}}^{(1)}$ of size $K-q$ in $\mathbf{G}$. Furthermore, the honest workers in $H$ form another clique $M_{\mathbf{G}}^{(2)}$ which is also of size $K-q$. Thus, the detection algorithm cannot select one over the other and the adversaries will evade detection; and the fallback robust aggregation strategy will apply.

\begin{table}[!h]
	\centering
	\caption{Parameters used for training.}
	\label{table:tuning}
	\begin{tabular}{P{0.9cm}P{1.3cm}P{3cm}}
		\hline
		Figure & Schemes & Learning rate schedule \\
		\hline
		\ref{fig:top1_fig_94} & 1,2,5,6 & $(0.01,0.7)$\\
		\ref{fig:top1_fig_94} & 3,4 & $(0.1,0.95)$\\
		\ref{fig:top1_fig_95} & 1 & $(0.001,0.95)$\\
		\ref{fig:top1_fig_96} & 1,2 & $(0.01,0.7)$\\
		\ref{fig:top1_fig_97} & 1,2 & $(0.1,0.7)$\\
		\ref{fig:top1_fig_97} & 3,4 & $(0.1,0.95)$\\
		\ref{fig:top1_fig_97} & 5,6 & $(0.01,0.7)$\\
		\ref{fig:top1_fig_98} & 1 & $(0.1,0.7)$\\
		\ref{fig:top1_fig_99} & 1,2 & $(0.01,0.975)$\\
		\ref{fig:top1_fig_101} & 1,2 & $(0.1,0.7)$\\
		\ref{fig:top1_fig_101} & 3,4 & $(0.1,0.95)$\\
		\ref{fig:top1_fig_101} & 5,6 & $(0.01,0.95)$\\
		\ref{fig:top1_fig_100} & 1 & $(0.1,0.95)$\\
		\ref{fig:top1_fig_100} & 2 & $(0.01,0.7)$\\
		\ref{fig:top1_fig_88} & 1 & $(0.01,0.7)$\\
		\ref{fig:top1_fig_88} & 2 & $(0.1,0.95)$\\
		\ref{fig:top1_fig_88} & 3 & $(0.01,0.7)$\\
		\ref{fig:top1_fig_102} & 1,2 & $(0.01,0.7)$\\
		\ref{fig:top1_fig_102} & 3 & $(0.1,0.95)$\\
		\ref{fig:top1_fig_103} & 2 & $(0.01,0.7)$\\
		\ref{fig:top1_fig_104} & 2 & $(0.01,0.95)$\\
		\hline
	\end{tabular}
\end{table}

\subsection{Experiment Setup Details}
\label{appendix:implementation_details}

\subsubsection{Cluster Setup}
\label{appendix:cluster_details}
We used clusters of $K=15$ and $21$ workers arranged in various setups within Amazon EC2. Initially, we used a PS of type \texttt{i3.16xlarge} and several workers of type \texttt{c5.4xlarge} to setup a distributed cluster. However, this requires training data to be transmitted from the PS to every single machine based on our current implementation; an alternative approach one can follow is to setup shared storage space accessible by all machines to store the training data. Also, some instances were automatically terminated by AWS per the AWS \emph{spot instance} policy limitations;\footnote{\href{https://docs.aws.amazon.com/AWSEC2/latest/UserGuide/spot-interruptions.html}{https://docs.aws.amazon.com/AWSEC2/latest/UserGuide/spot-interruptions.html}} this incurred some delays in resuming the experiments that were stopped. In order to facilitate our evaluation and to avoid these issues we decided to simulate the PS and the workers for the rest of the experiments on a single instance of type \texttt{x1.16xlarge}. We emphasize that the choice of the EC2 setup does not affect any of the numerical results in this paper since in all cases we used a single virtual machine image with the same dependencies.

\subsubsection{Dataset Preprocessing and Hyperparameter Tuning}
The images have been normalized using standard values of mean and standard deviation for the dataset. The value used for momentum (for gradient descent) was set to $0.9$ and we trained for $16$ epochs in all experiments. The number of epochs is precisely the invariant we maintain across all experiments, i.e., all schemes process the training data the same number of times. The batch size and the learning rate are chosen independently for each method; the number of iterations are adjusted accordingly to account for the number of epochs. We followed the advice of the authors of DETOX and chose $(K, b)=(15, 480)$ and $(K, b)=(21, 672)$ for the DETOX and baseline schemes. For Aspis, we used $(K, b)=(15, 14560)$ (32 samples per file) and $(K, b)=(21, 3990)$ (3 samples per file) for the ALIE experiments and $b=1365$ (3 samples per file) for the remaining experiments except for the FoE optimal attack $q=4$ (\emph{cf.} Figure \ref{fig:top1_fig_101}) for which $b=14560$ performed better. In Table \ref{table:tuning}, a learning rate schedule is denoted by $(x,y)$; this notation signifies the fact that we start with a rate equal to $x$ and every $z$ iterations we set the rate equal to $x\times y^{t/z}$, where $t$ is the index of current iteration and $z$ is set to be the number of iterations occurring between two consecutive checkpoints in which we store the model (points in the accuracy figures). We will also index the schemes in order of appearance in the corresponding figure's legend. Experiments which appear in multiple figures are not repeated in Table \ref{table:tuning} (we ran those training processes once). In order to pick the optimal hyperparameters for each scheme, we performed an extensive grid search involving different combinations of $(x,y)$. In particular, the values of $x$ we tested are 0.1, 0.01 and 0.001 and for $y$ we tried 1, 0.975, 0.95, 0.7 and 0.5. For each method, we ran 3 epochs for each such combination and chose the one which was giving the lowest value of average cross-entropy loss (principal criterion) and the highest value of top-1 accuracy (secondary criterion).

\begin{figure*}[!ht]
	\centering
	\begin{subfigure}[b]{0.43\textwidth}
		\centering
		\includegraphics[scale=0.45]{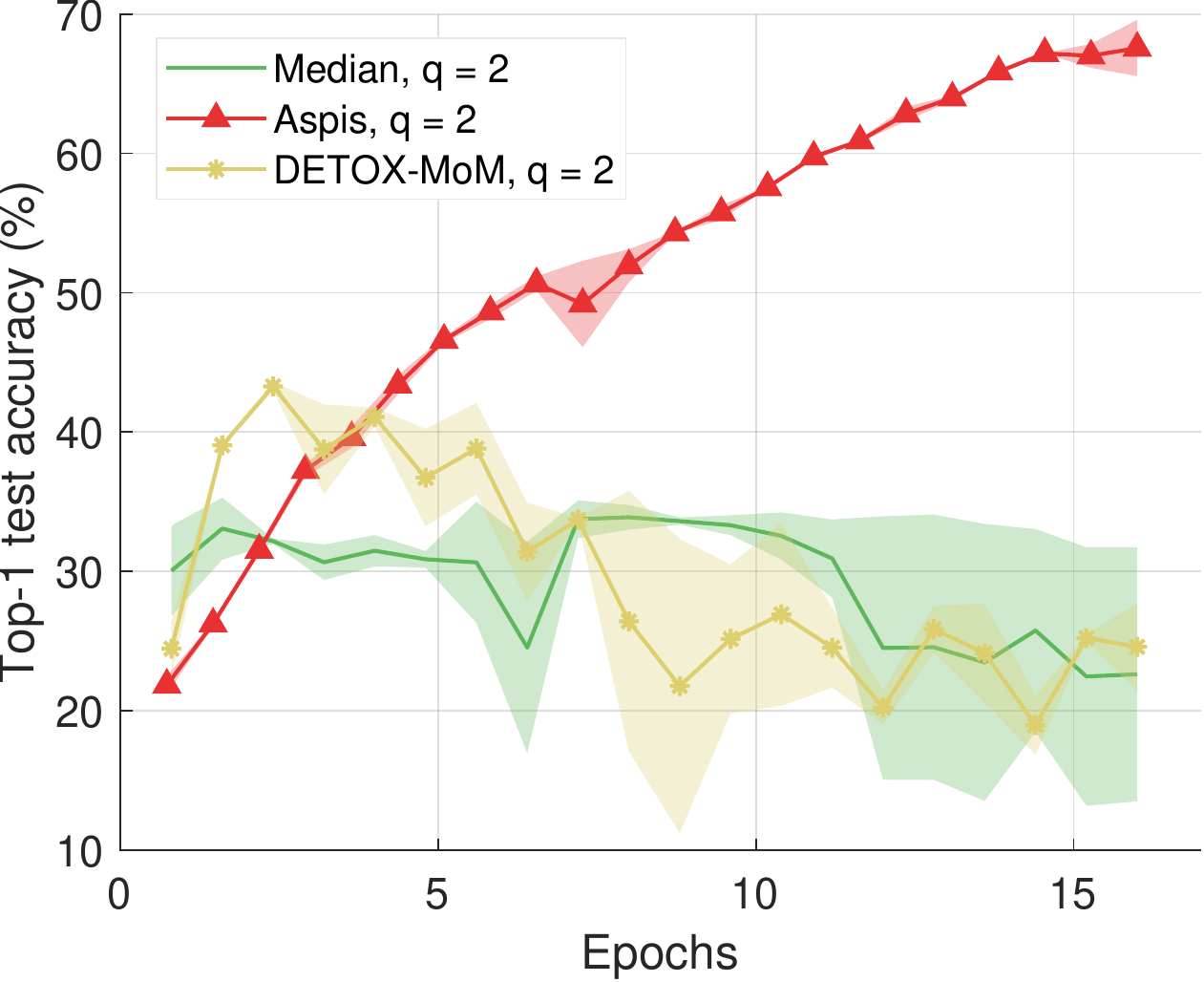}
		\caption{$q=2$ adversaries.}
		\label{fig:top1_error_bar_1}
	\end{subfigure}
	\hspace{0.07\textwidth}
	\begin{subfigure}[b]{0.43\textwidth}
		\centering
		\includegraphics[scale=0.45]{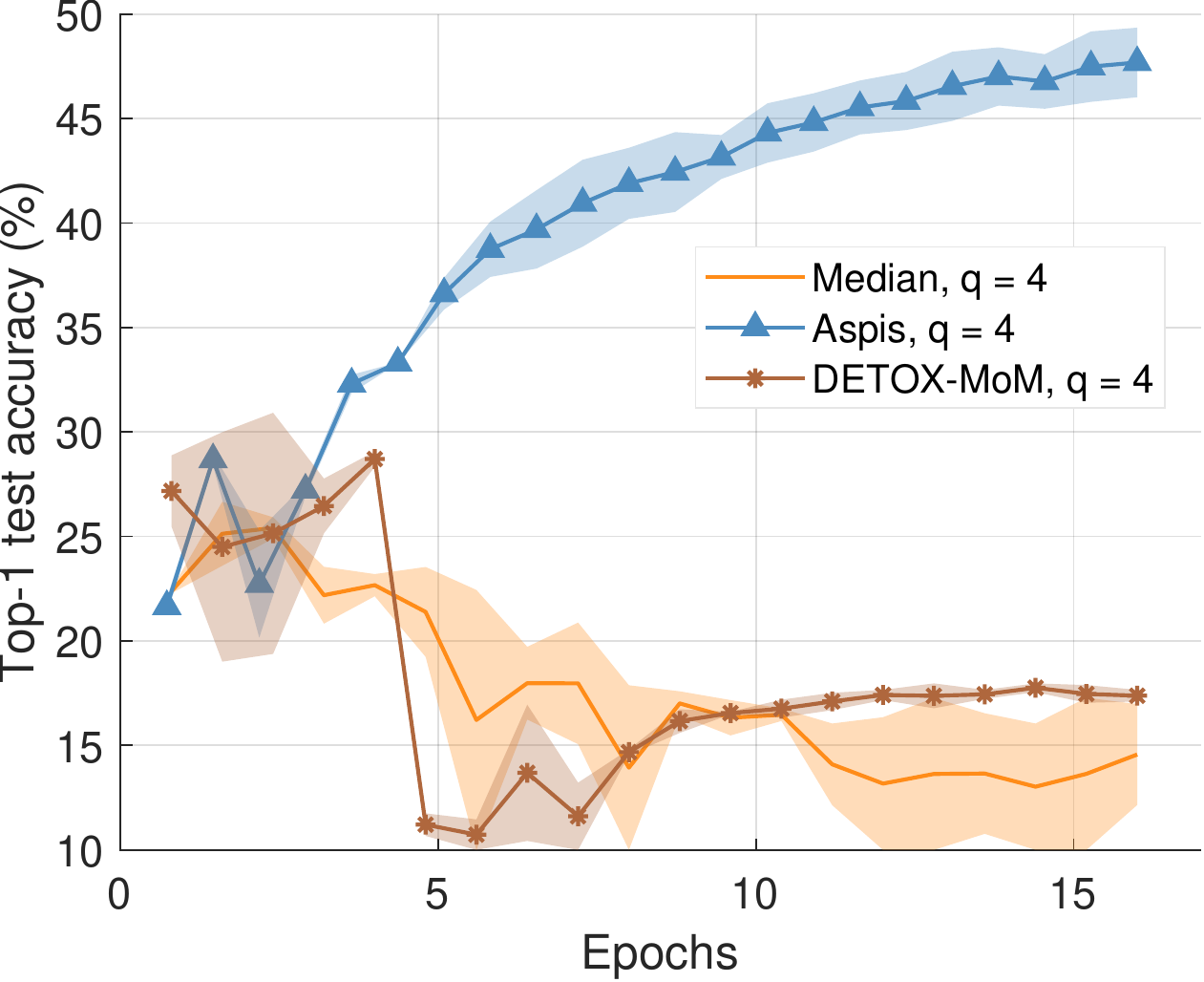}
		\caption{$q=4$ adversaries.}
		\label{fig:top1_error_bar_4}
	\end{subfigure}
	\caption{\emph{ALIE} optimal attack and median-based defenses (CIFAR-10), $K=15$ with different random seeds.}
	\label{fig:top1_error_bar_median}
\end{figure*}

\begin{figure*}[!ht]
	\centering
	\begin{subfigure}[b]{0.43\textwidth}
		\centering
		\includegraphics[scale=0.45]{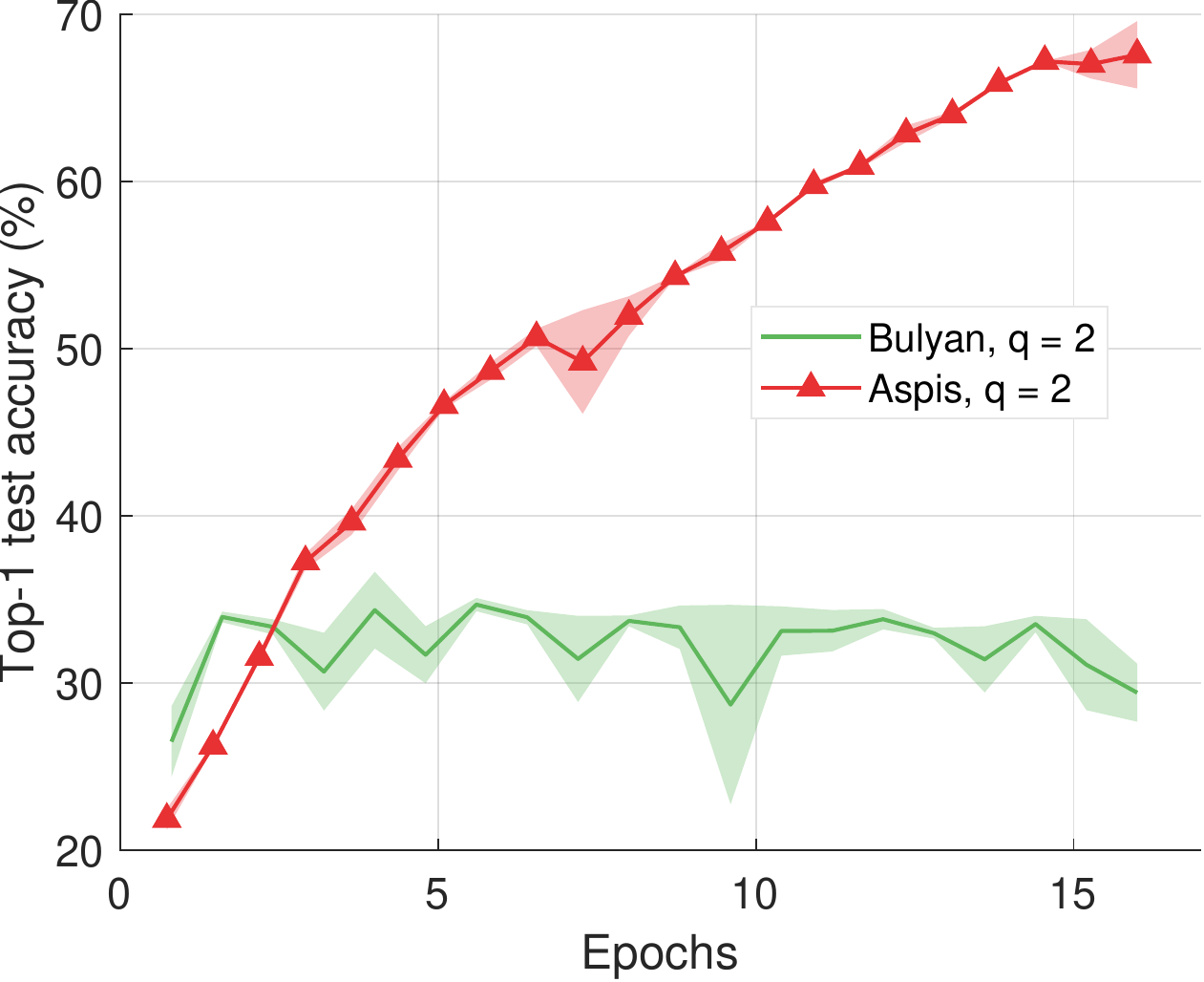}
		\caption{$q=2$ adversaries.}
		\label{fig:top1_error_bar_2}
	\end{subfigure}
	\hspace{0.07\textwidth}
	\begin{subfigure}[b]{0.43\textwidth}
		\centering
		\includegraphics[scale=0.45]{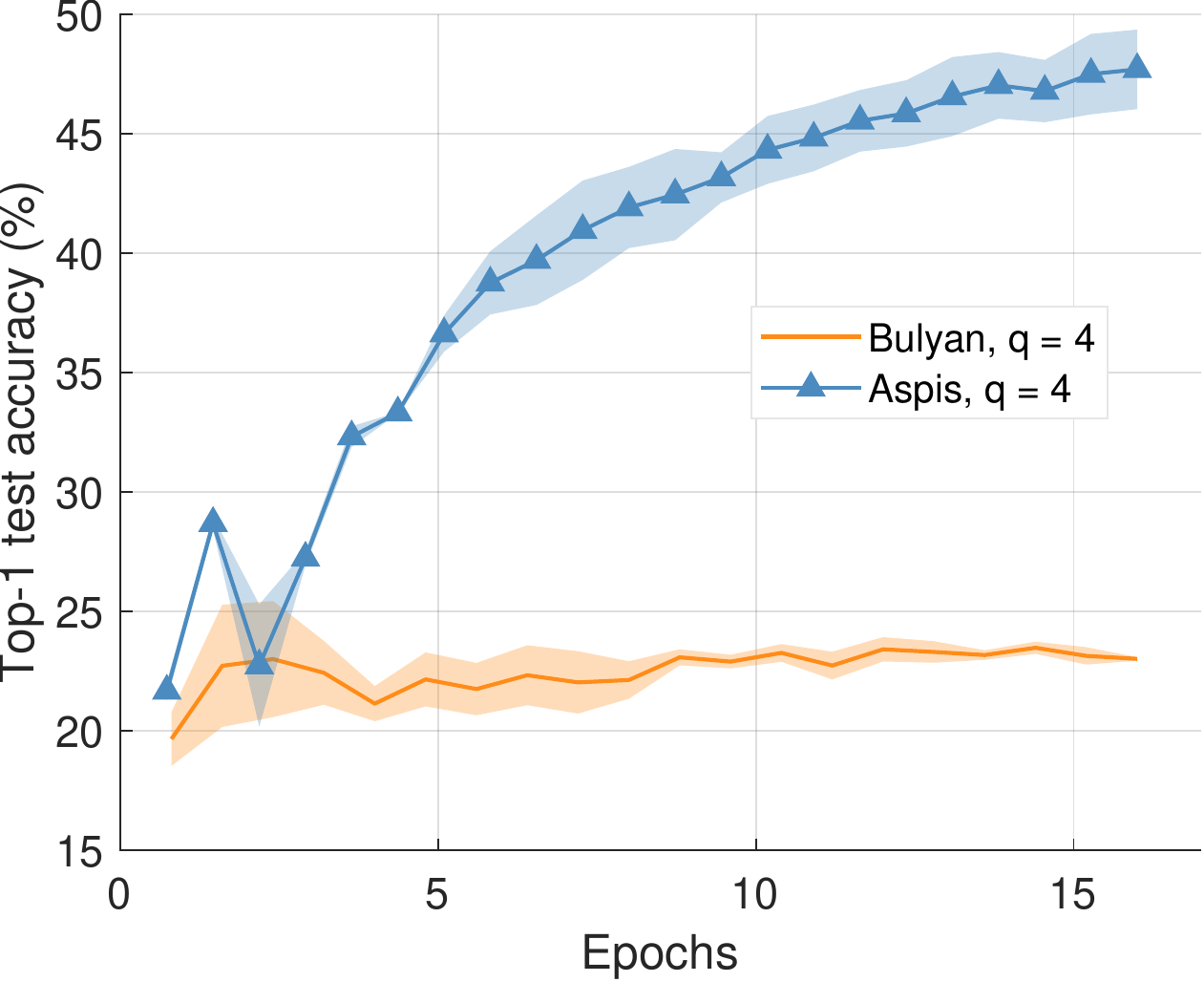}
		\caption{$q=4$ adversaries.}
		\label{fig:top1_error_bar_5}
	\end{subfigure}
	\caption{\emph{ALIE} optimal attack and \emph{Bulyan}-based defenses (CIFAR-10), $K=15$ with different random seeds.}
	\label{fig:top1_error_bar_bulyan}
\end{figure*}

\begin{figure*}[!ht]
	\centering
	\begin{subfigure}[b]{0.43\textwidth}
		\centering
		\includegraphics[scale=0.45]{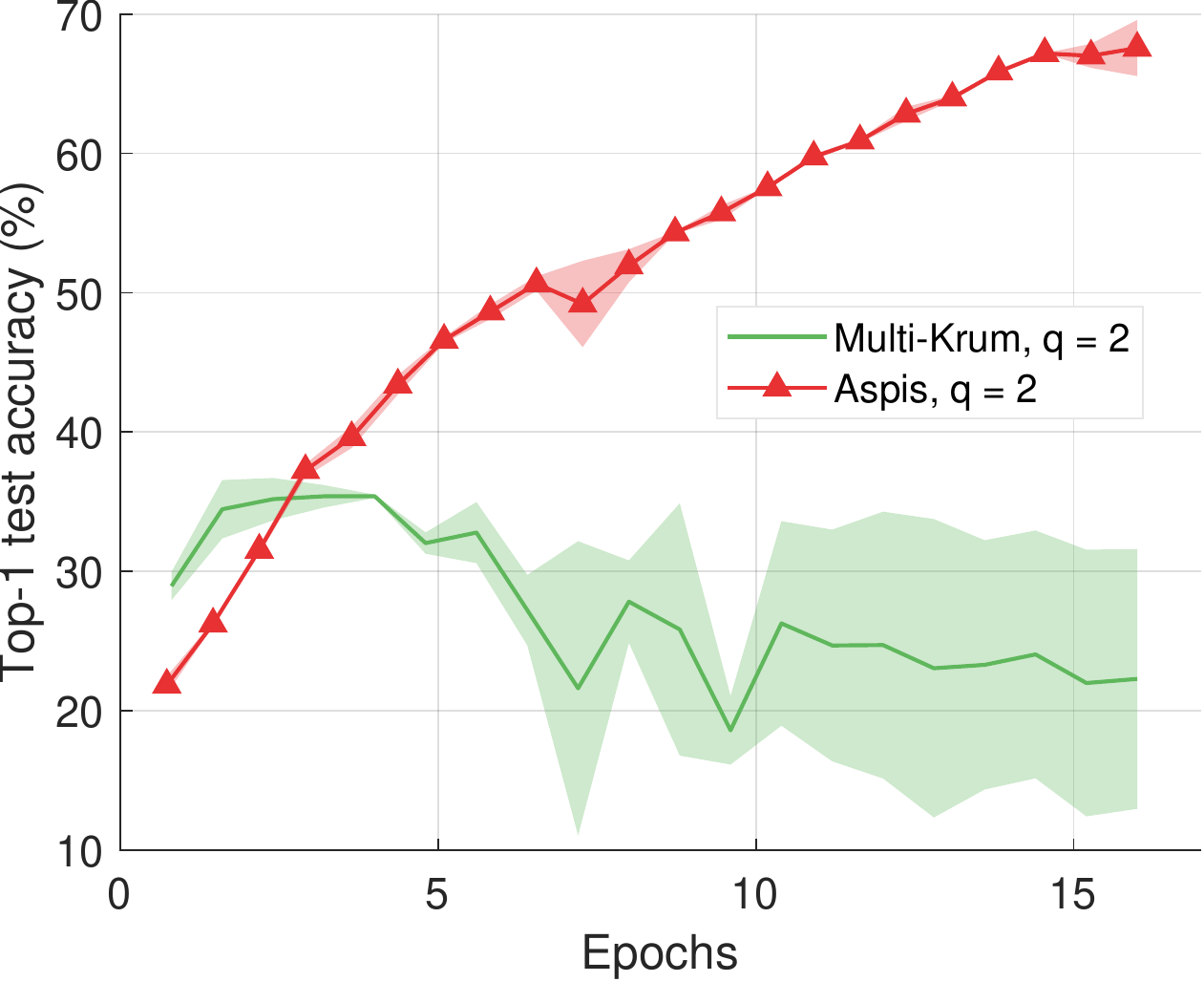}
		\caption{$q=2$ adversaries.}
		\label{fig:top1_error_bar_3}
	\end{subfigure}
	\hspace{0.07\textwidth}
	\begin{subfigure}[b]{0.43\textwidth}
		\centering
		\includegraphics[scale=0.45]{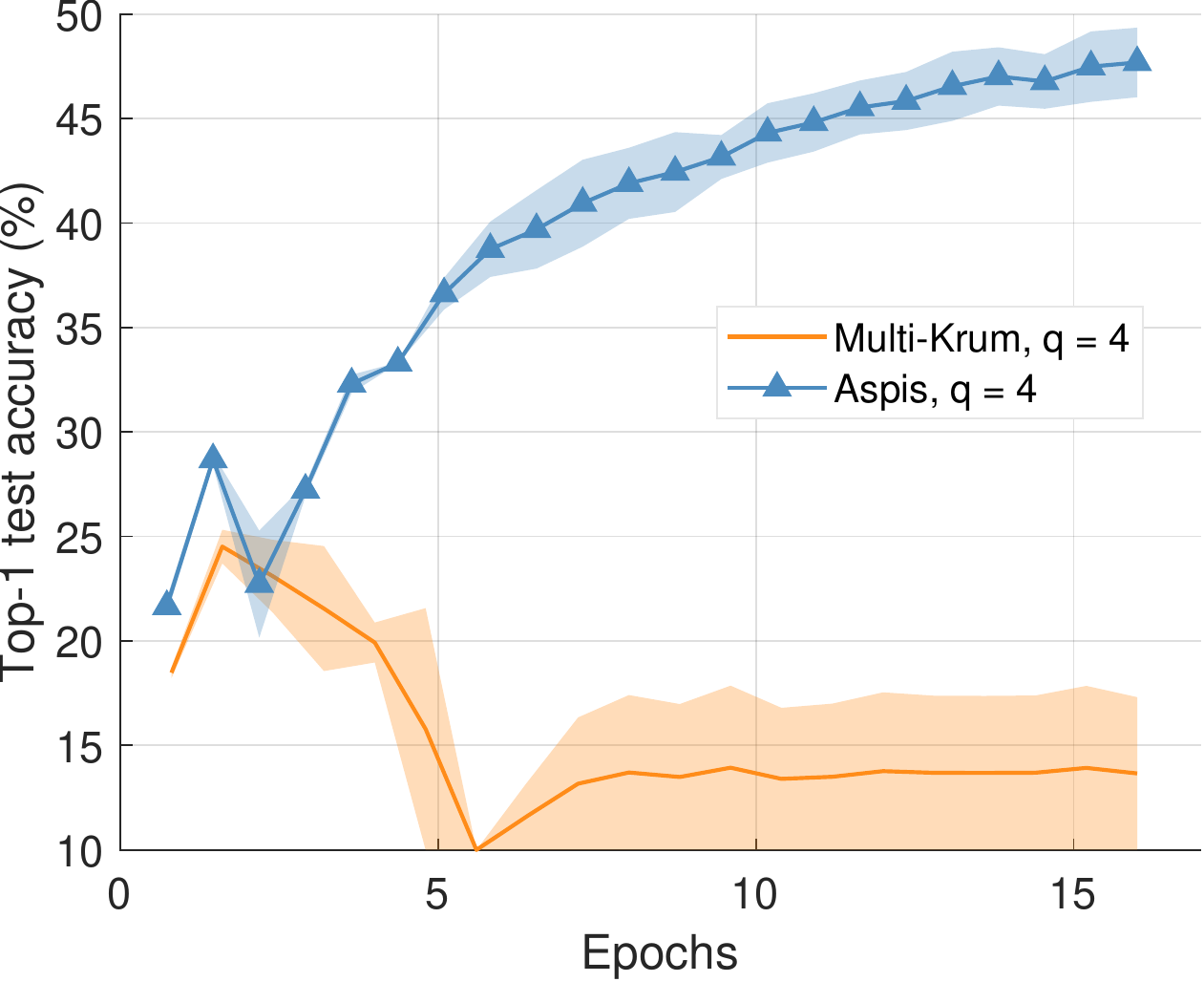}
		\caption{$q=4$ adversaries.}
		\label{fig:top1_error_bar_6}
	\end{subfigure}
	\caption{\emph{ALIE} optimal attack and \emph{Multi-Krum}-based defenses (CIFAR-10), $K=15$ with different random seeds.}
	\label{fig:top1_error_bar_multikrum}
\end{figure*}

\subsubsection{Error Bars}
\label{appendix:error_bars}
In order to examine whether the choice of the random seed affects the accuracy of the trained model we have performed the experiments for the ALIE distortion for two different seeds for the values $q=2,4$ for every scheme; we used $428$ and $50$ as random seeds. These tests have been performed for the case of $K=15$ workers. In Figure \ref{fig:top1_error_bar_1}, for a given method we report the minimum accuracy, the maximum accuracy and their average for each evaluation point. We repeat the same process in Figures \ref{fig:top1_error_bar_2} and \ref{fig:top1_error_bar_3} when comparing with Bulyan and Multi-Krum, respectively. The corresponding experiments for $q=4$ are shown in the Figures \ref{fig:top1_error_bar_4}, \ref{fig:top1_error_bar_5} and \ref{fig:top1_error_bar_6}.

Given the fact that these experiments take a significant amount of time and that they are computationally expensive, we chose to perform this consistency check for a subset of our experiments. Nevertheless, these results indicate that prior schemes \cite{detox, aggregathor, bulyan} are very sensitive to the choice of the random seed and demonstrate an unstable behavior in terms of convergence. In all of these cases the achieved value of accuracy at the end of the 16 epochs of training is small compared to Aspis. On the other hand, the accuracy results for Aspis are almost identical for both choices of the random seed.

\subsubsection{Computation and Communication Overhead}
\label{appendix:computation_communication_overhead}
Our scheme provides robustness under powerful attacks and sophisticated distortion methods at the expense of increased computation and communication time. Note that each worker has to perform $l$ forward/backward propagation computations and transmit $l$ gradients per iteration. In related baseline \cite{bulyan,blanchard_krum} and redundancy-based methods \cite{detox, draco} each worker is responsible for a single such computation. Experimentally, we have observed that Aspis needs up to $5\times$ overall training time compared to other schemes to complete the same number of training epochs. We emphasize that the training time incurred by each scheme depends on a wide range of parameters including the utilized defense, the batch size and the number of iterations and can vary significantly. We believe that implementation-related improvements such as utilizing GPUs can alleviate some of the overhead. Communication-related algorithmic improvements are also worth exploring. Finally, our implementation natively supports resuming from a checkpoint (trained model) and hence, when new data becomes available we can only use that data to perform more epochs of training. 
\subsubsection{Software}
\label{appendix:software}
Our implementation of the Aspis algorithm used for the experiments builds on the ByzShield's \mbox{\cite{byzshield}} PyTorch skeleton 
and has been provided along with dependency information and instructions \footnote{\href{https://www.dropbox.com/sh/8bf4itlb3tz8o4n/AABKp6PaGlr_M8tRVmIRM6Pba?dl=0}{https://www.dropbox.com/sh/8bf4itlb3tz8o4n/AABKp6PaGlr\_M8tRVmIRM6Pba?dl=0}}. The implementation of ByzShield is available at \href{https://github.com/kkonstantinidis/ByzShield}{https://github.com/kkonstantinidis/ByzShield} and uses the standard Github license. We utilized the NetworkX package \cite{networkx} for the clique-finding; its license is 3-clause BSD. The CIFAR-10 dataset \cite{cifar10} comes with the MIT license; we have cited its technical report, as required.

\begin{table}[!ht]
	\centering
	\caption{Main notation of the paper.}
	\label{table:notation}
	\begin{tabular}{P{1cm}p{6cm}}
		\hline
		Symbol & Meaning \\
		\hline
		$K$ & number of workers\\
		$q$ & number of adversaries\\
		$r$ & redundancy (number of workers each file is assigned to)\\
		$b$ & batchsize\\
		$B_t$ & samples of batch of $t^{\mathrm{th}}$ iteration\\
		$f$ & number of files (alternatively called \emph{groups} or \emph{tasks})\\
		$U_j$ & $j^{\mathrm{th}}$ worker\\
		$l$ & computation load (number of files per worker)\\
		$\calN(U_j)$ & set of files of worker $U_j$\\
		$\calN(B_{t,i})$ & set of workers assigned to file $B_{t,i}$\\
		$\mathbf{g}_{t,i}$ & true gradient of file $B_{t,i}$ with respect to $\mathbf{w}$\\
		$\hat{\mathbf{g}}_{t,i}^{(j)}$ & returned gradient of $U_j$ for file $B_{t,i}$ with respect to $\mathbf*{w}$\\
		$\mathbf{m}_i$ & majority gradient for file $B_{t,i}$\\
		$\calU$ & worker set $\{U_1,U_2,...,U_K\}$\\
		$\mathbf{G}$ & graph indicating the agreements of pairs of workers in all of their common gradient tasks\\
		$A$ & set of adversaries\\
		$M_{\mathbf{G}}$ & maximum clique in $\mathbf{G}$\\
		$c^{(q)}$ & number of distorted gradients after detection and aggregation\\
		$c_{\mathrm{max}}^{(q)}$ & maximum number of distorted gradients after detection and aggregation (worst-case)\\
		$D_i$ & disagreement set (of workers) for $i^{\mathrm{th}}$ adversary\\
		$r'$ & $(r+1)/2$, i.e., minimum number of distorted copies needed to corrupt majority vote for a file\\
		$\epsilon$ & $c^{(q)}/f$, i.e., fraction of distorted gradients after detection and aggregation\\
		$X_j$ & subset of files where the set of active adversaries is of size $j$\\
		
		\hline
	\end{tabular}
\end{table}


\end{document}